\PassOptionsToPackage{sort&compress}{natbib}
\documentclass[]{fairmeta}

\usepackage{amsmath}
\usepackage{amssymb}
\usepackage{float}
\usepackage{makecell}

\setcitestyle{numbers,square,comma}
\newcommand{\fitbox}[2]{\setbox0=\hbox{#2}\ifdim\wd0>#1\relax\resizebox{#1}{!}{\usebox0}\else\usebox0\fi}

\newcommand{\dsTwoV}{OCT-Cam-2V}
\newcommand{\dsFourV}{OCT-Cam-4V}
\newcommand{\dsNymeria}{Nymeria}
\newcommand{\dsMasq}{LightDome}

\title{EyeDiT: Anatomy-Grounded Multi-Modal Eye Generation and Simulation}

\author[1*,2]{Yonghan~Lee}
\author[1]{Zachary~Willms}
\author[1]{Kun~Liu}
\author[1]{Pushkar~Anand}
\author[1]{Ali~Behrooz}
\author[1]{\newline Dmitri~Model}
\author[2]{Dinesh~Manocha}
\author[1]{Yatong~An}

\affiliation[1]{Meta Reality Labs}
\affiliation[2]{University of Maryland, College Park}

\contribution[*]{Work done during an internship at Meta}

\abstract{
Synthetic eye data are critical for rapid development of eye tracking (ET) in AR/VR applications. However, existing synthetic eye generators are largely image-centric and lack a high-quality underlying 3D anatomy of the eye,
which limits their control mainly to the gaze and provides no anatomical labels.
To address this problem, we propose EyeDiT, an anatomy-grounded, multi-modal 3D eye generation and simulation model trained with optical coherence tomography (OCT), labeled anatomy point clouds and near-eye images.
Our pipeline first trains a Multi-Modal Anatomy Variational Autoencoder (VAE), which encodes an anatomy-aware latent representation from all input modalities. Then, our Eye Latent DiT is trained to generate latent samples conditioned on the optical axis, the pupil position and radius, and the subject identity. Together with Gaussian decoders for novel-view simulation, our model enables control over the pupil, subject identity and viewpoint for data synthesis.
We show that our model outperforms previous eye image synthesis models in novel-view synthesis (24.9 against 21.8\,dB PSNR from a single view), and that a public ET neural network trained on our synthetic data shows a smaller sim-to-real gap than when trained on data from other generators (3.5$^\circ$ vs.\ 8.4$^\circ$ gaze error), approaches a network trained on real data, and further improves ET performance when combined with real data (2.3$^\circ$ vs.\ 2.6$^\circ$). More information is available on the project page: \url{https://lyhan12.github.io/eyedit}.
}

\metadata[Keywords]{Synthetic Eye Data Generation, Eye Tracking, Multi-Modal 3D Generation, Sim-to-Real}
\correspondence{Yatong An at \email{yatong@meta.com}}

\begin{document}

\maketitle
\suppressfloats[t]

\begin{figure}[t]
    \centering
    \includegraphics[width=\linewidth]{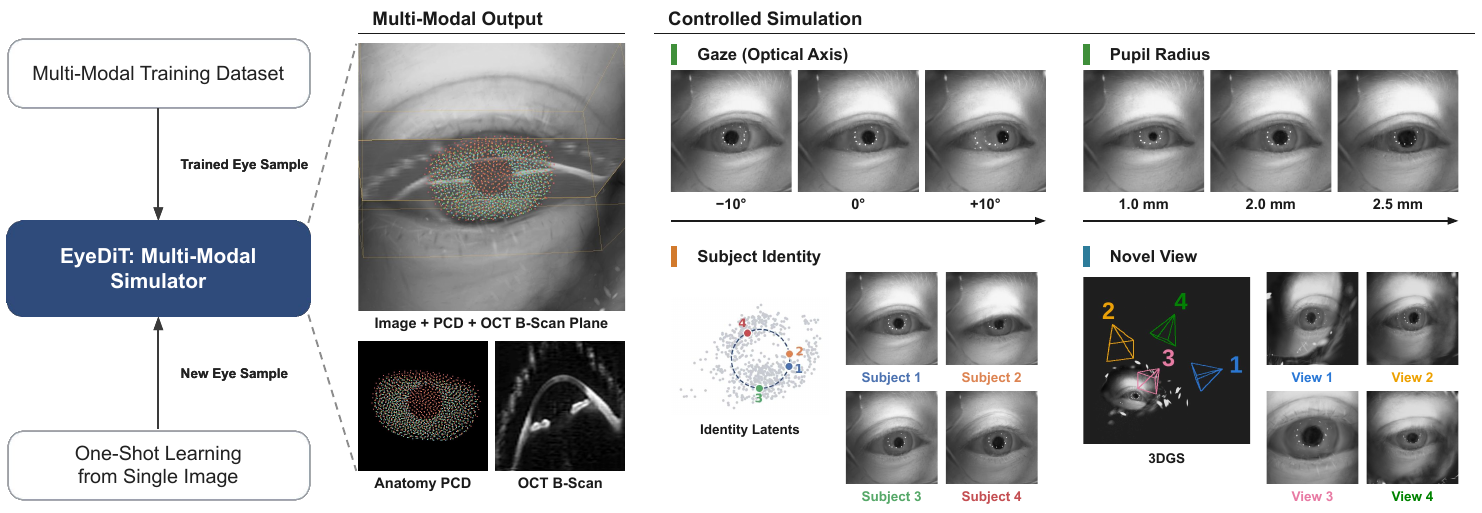}
    \caption{We present EyeDiT, which demonstrates controlled simulation (right) of gaze, pupil radius, subject identity and novel views for synthetic near-eye images. The examples show simulated gaze and pupil radius sweeps, identities from the learned identity embedding, and novel views rendered from decoded 3D Gaussians~\cite{Kerbl2023GS}. Our core novelty is anatomy-grounded eye data generation (middle), in which images, anatomy-labeled (iris and cornea) point clouds and OCT B-scans are learned and generated jointly. With this anatomical grounding, our method accurately controls both gaze and pupil radius, whereas previous learned generators mainly control the gaze~\cite{Li2025ControllableGaze, Dumery2026GazePrior, Wei2025GazeGaussian, Li2022EyeNeRF}. The other key capability (left) is one-shot reconstruction and simulation of a new eye, together with simulation of the trained eyes.}
    \label{fig:intro}
\end{figure}

\section{Introduction}

\label{sec:intro}

\begin{table*}[t]
    \centering
    \caption{Capability comparison of eye modeling methods. Our EyeDiT learns and generates multi-modal outputs and controls the pupil radius. Our formulation also renders novel views and reconstructs new eyes from a single image. GazeGaussian and GazePrior handle new subjects with test-time fit on their images (ft: fine-tuning, opt: optimization), while our model needs only a single forward pass. NVGaze and UnityEyes 2 use pre-defined mesh models, and have limited modeling capability for real subjects.}
    \label{tab:capability_comparison}
    \fitbox{0.95\textwidth}{%
    \begin{tabular}{llcccccccc}
        \toprule
        & & \multicolumn{3}{c}{\textbf{Multi-Modal Output}}
        & \multicolumn{2}{c}{\textbf{Pupil Control}}
        & \multicolumn{2}{c}{\textbf{Subject Simulation}}
        & \multirow{2}{*}{\makecell{\textbf{Novel View}\\\textbf{Rendering}}} \\
        \cmidrule(lr){3-5}
        \cmidrule(lr){6-7}
        \cmidrule(lr){8-9}
        \textbf{Method} & \textbf{Type}
        & Image & Anatomy (PCD) & OCT
        & Gaze & Radius
        & Learned Subj. & New Subj.
        & \\
        \midrule
        NVGaze~\cite{Kim2019NVGaze} & Parametric & \checkmark & $\times$ & $\times$ & \checkmark & \checkmark & $\times$ & $\times$ & Mesh \\
        UnityEyes 2~\cite{Smith2025UnityEyes2} & Parametric & \checkmark & $\times$ & $\times$ & \checkmark & \checkmark & $\times$ & $\times$ & Mesh \\
        Lin et al.~\cite{Lin2026EyeTracker} & Per-subject & \checkmark & $\times$ & $\times$ & $\times$ & $\times$ & per-subject & per-subject fit & Mesh+NeRF \\
        ControllableGaze~\cite{Li2025ControllableGaze} & Per-subject & \checkmark & $\times$ & $\times$ & \checkmark & $\times$ & per-subject & per-subject fit & 3DGS \\
        GazeGaussian~\cite{Wei2025GazeGaussian} & Learned & \checkmark & $\times$ & $\times$ & \checkmark & $\times$ & shared model & one-shot (ft) & 3DGS \\
        GazePrior~\cite{Dumery2026GazePrior} & Learned & \checkmark & $\times$ & $\times$ & \checkmark & $\times$ & latent prior & few-shot (opt+ft) & NeRF \\
        \midrule
        \textbf{Ours} & Learned & \checkmark & \checkmark & \checkmark & \checkmark & \checkmark & latent prior & one-shot (feed-forward) & 3DGS \\
        \bottomrule
    \end{tabular}%
    }
\end{table*}

AR/VR products usually rely on eye tracking (ET)~\cite{Engel2023ProjectAria,Mansour2025AriaET,Lin2026EyeTracker,Shapira2026GazeShift} to estimate the status of the eye, including gaze direction, pupil position and pupil radius. These capabilities are critical to enable downstream functions in AR/VR applications such as foveated rendering~\cite{Guenter2012Foveated,Patney2016Foveated,Deng2022FoVNeRF,Franke2025VRSplatting} and gaze-based interaction~\cite{Pfeuffer2017GazePinch,Sidenmark2019EyeHead,Plopski2022EyeXR}. However, developing a reliable eye tracker typically requires large-scale near-eye datasets with ground-truth annotations spanning diverse gaze directions, pupil sizes, and subjects, whose collection is costly and labor-intensive~\cite{Lin2026EyeTracker, Dumery2026GazePrior}. Moreover, each new hardware prototype with a different camera configuration necessitates a new round of data collection. Synthetic eye data generation offers a scalable alternative, enabling efficient development and validation of ET models on new AR/VR prototypes~\cite{Lin2026EyeTracker} with low cost and quick turnaround.

Yet, existing near-eye datasets and synthetic eye generators are largely image-centric: they provide images with gaze labels or 2D annotations~\cite{Dumery2026GazePrior,Li2025ControllableGaze}, without modeling the 3D anatomy (iris and cornea) behind them. As a result, their control is limited to gaze directions~\cite{Dumery2026GazePrior,Li2025ControllableGaze,Wei2025GazeGaussian}, and they cannot provide readouts of the underlying anatomy, such as the corneal surface, pupil position and radius, or optical axis. Recovering this anatomy from images alone is difficult for two reasons. First, the cornea is transparent and refractive, so its curvature and the true shape of the iris are not directly visible in images. Second, AR/VR devices observe the eye through only a few cameras at fixed positions, which leaves its 3D structure severely under-constrained.

One way to measure this anatomy directly is optical coherence tomography (OCT), which is widely used in clinical ophthalmology~\cite{Huang1991OCT,Shan2019AnteriorOCTReview}. Similar to a scanning LiDAR, OCT measures depth with light and sweeps its beam horizontally and vertically across the eye, building an accurate 3D volume of the cornea, iris, and pupil~\cite{Qian2026WEOCT}. However, OCT devices are typically operated in controlled settings and cannot be used in practical scenarios with smart glasses or headsets with near-eye cameras. In addition, anterior-segment OCT covers only the region around the cornea~\cite{Shan2019AnteriorOCTReview,Recchioni2024CASIA2}, and does not capture the periocular region of the eye. As a result, OCT has rarely been combined with near-eye images in typical AR/VR scenarios.

Meanwhile, recent advancements in generative models~\cite{Rombach2022LDM,Peebles2023DiT,Zhang2024CLAY,Xiang2025TRELLIS} have shown that large data-driven priors can be learned offline and then used to generate structurally consistent objects from only a few observations at test time. In eye image synthesis tasks, GazePrior~\cite{Dumery2026GazePrior} adopts a similar idea. Because all eyes share a common global structure, it learns this structure as a prior and adapts only the subject-specific variation from a few views. However, its prior is learned from images alone and therefore does not capture the internal anatomy of the eye, such as the cornea and iris, which images cannot reveal.

In this regard, our key idea is to use OCT and anatomical knowledge at training time to learn a multi-modal prior for near-eye data generation. At test time, this prior supports both controllable simulation and one-shot simulation of new subjects from context images, without retraining.

\begin{figure*}[!t]
    \centering
    \includegraphics[width=\textwidth]{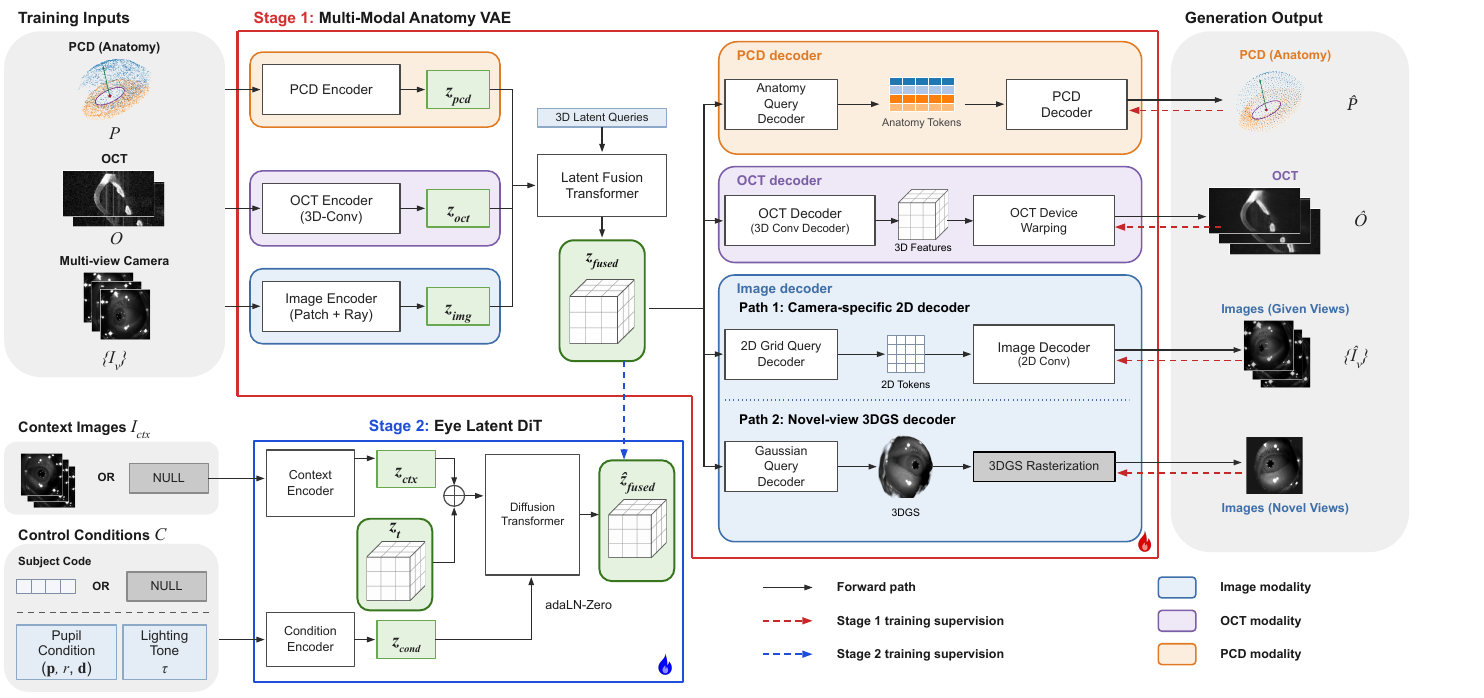}
    \caption{Overview of EyeDiT. (Stage 1) Our multi-modal anatomy VAE learns a shared latent space from OCT volumes, near-eye images and anatomy point clouds. (Stage 2) Our eye latent DiT is then trained on this latent space to simulate various gaze, pupil, and identity conditions. Our pipeline generates multi-modal outputs for a learned identity or for a new subject given by a context image.}
    \label{fig:architecture}
\end{figure*}

Building on this idea, we propose EyeDiT, an anatomy-grounded, multi-modal 3D eye generation and simulation model (Figs.~\ref{fig:intro} and~\ref{fig:architecture}). EyeDiT first learns a shared latent space of near-eye images, OCT volumes, and anatomy point clouds with a variational autoencoder, so that each latent sample describes both the eye anatomy and its periocular appearance. A diffusion transformer is trained to generate samples in this space, conditioned on the gaze (optical axis), the pupil radius and subject identity, and a Gaussian decoder renders the generated eye from novel views.

\noindent \textbf{Main Results:} To the best of our knowledge, EyeDiT is the first eye generation and simulation model grounded in anatomy, using OCT B-scans and derived anatomy point clouds. Tab.~\ref{tab:capability_comparison} compares its capabilities with those of prior eye modeling methods. Our main contributions are:

\begin{itemize}
\item \textbf{Anatomy-grounded multi-modal eye generation.} EyeDiT generates near-eye images, along with underlying OCT volumes and anatomy point clouds for iris and cornea. This anatomical output enables readout of diverse biometrics, including the optical axis, pupil position and size, and corneal surface. We report the reconstruction quality and its biometrics in Tab.~\ref{tab:multi_modal_recon} and Fig.~\ref{fig:multi_modal_recon}.
\item \textbf{Simulation capabilities over pupil, identity, and views.} Beyond gaze-only simulation, EyeDiT simulates (1) the gaze and the pupil radius (Fig.~\ref{fig:qual_condition_sweep}), (2) subject identity for both trained and new eyes, and (3) novel views through the Gaussian decoder (Tab.~\ref{tab:nv_merged}). This enables controlled generation of data that are hard to collect in real settings.
    \item \textbf{Synthetic data for downstream ET methods.} We train the Aria ET estimator~\cite{Mansour2025AriaET} on data from different eye generators~\cite{Dumery2026GazePrior,Li2025ControllableGaze, Smith2025UnityEyes2}. Training on our data gives the lowest errors among the generators and approaches the accuracy of training on real data, and augmenting the real data with ours further improves it (Tab.~\ref{tab:sim_to_real_aria}).
\end{itemize}

\section{Related Work}
\label{sec:related_works}

\noindent \textbf{Eye Datasets and Eye Tracking} Training eye-tracking methods often requires large near-eye image datasets~\cite{Engel2023ProjectAria,Mansour2025AriaET,Garbin2020OpenEDS,Shapira2026GazeShift}. Since such datasets are expensive to collect, many synthetic eye image generation methods have been studied as an alternative~\cite{Wood2016UnityEyes,Kim2019NVGaze,Smith2025UnityEyes2,Lin2026EyeTracker,Dumery2026GazePrior,Li2025ControllableGaze,Wei2025GazeGaussian}. Most of them, however, are image-centric and lack anatomical awareness. On the other hand, OCT has been widely used in clinical settings~\cite{Huang1991OCT,Shan2019AnteriorOCTReview} for precise 3D measurement of the anterior segment and the whole eye~\cite{Qian2026WEOCT,Ma2025MCOA}. Yet combining it with anatomically consistent image synthesis remains less explored.

\noindent \textbf{Latent 3D Generation Models} Latent diffusion~\cite{Kingma2014VAE,Rombach2022LDM,Peebles2023DiT,Ho2022CFG} trained on large 3D asset collections yields strong priors for 3D generation and editing~\cite{Zhang20233DShape2VecSet,Zhang2024CLAY,Xiang2025TRELLIS,Xiang2026TRELLIS2,Lan2025GaussianAnything}. A single latent can also hold several modalities~\cite{Krishnan2025Orchid,Shi2019MMVAE,Tang2023CoDi}, with spatial conditions attached to it~\cite{Zhang2023ControlNet,Li2024ControlNetPP}. Following this design, we build a latent 3D generation model whose latent is shared by near-eye images, OCT volumes and anatomy point clouds.

\noindent \textbf{Novel-View Synthesis} NeRF~\cite{Mildenhall2020NeRF,Barron2021MipNeRF,Muller2022InstantNGP} and 3D Gaussian splatting~\cite{Kerbl2023GS,Huang20242DGS,Yu2024MipSplatting,Wu20253DGUT} learn implicit and explicit scene representations for novel-view synthesis. They usually require per-scene optimization, but recent feed-forward models infer 3D Gaussians directly from a few context images~\cite{Charatan2024pixelSplat,Chen2024MVSplat,Zhang2024GSLRM,Xu2025DepthSplat,Ye2026YoNoSplat,Xu2026ReSplat,Ren2026TokenGS}, and geometry and video foundation models offer general scene priors~\cite{Wang2025MoGe2,Yang2024DepthAnythingV2,Keetha2026MapAnything,Yu2025ViewCrafter}. For the eye, 3D rendering models still optimize or fine-tune on each subject or capture, with NeRF~\cite{Li2022EyeNeRF,Lin2026EyeTracker,Dumery2026GazePrior} or Gaussian head models~\cite{Li2025ControllableGaze,Wei2025GazeGaussian}. Instead, we predict 3D Gaussians in a single forward pass with a Gaussian decoder attached to the eye latent.

\section{Methods}

\begin{figure*}[!t]
    \centering
    \includegraphics[width=\textwidth]{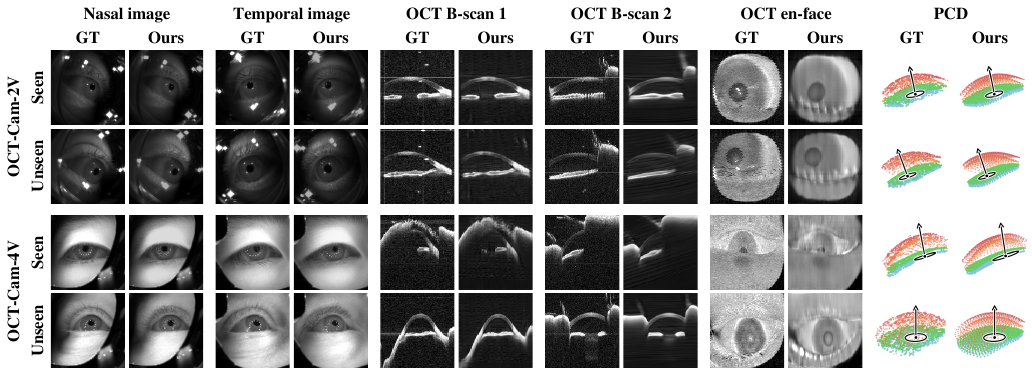}
    \caption{Inputs (GT) and reconstructions (Ours) of our multi-modal anatomy VAE on \dsTwoV{} (top two rows) and \dsFourV{} (bottom two rows), showing that our VAE preserves the main structures and details of the given inputs across all modalities. The columns show two views (nasal, temporal), two OCT B-scan positions, the projected OCT volume (en-face) and the PCD, each as a GT and Ours pair. Each row represents a subject either seen or unseen during training. Note that our VAE also reconstructs the unseen subjects faithfully.}
    \label{fig:multi_modal_recon}
\end{figure*}

\noindent \textbf{Problem Setting.} Our goal is to learn a controllable eye simulator that generates near-eye images $\hat{I}_v$ at any camera view $v$, an OCT volume $\hat{O}$ of the anterior eye anatomy (iris, cornea), and a labeled anatomy point cloud $\hat{P}$ with four labels, the front and back surfaces of the cornea and the iris (Fig.~\ref{fig:multi_modal_recon}). The outputs are controlled by the conditions $\mathcal{C}=\{\mathbf{p}, r, \mathbf{d}, \tau, s\}$: the pupil position $\mathbf{p}\in\mathbb{R}^3$ (mm), the pupil radius $r$ (mm), the optical axis (OA) $\mathbf{d}$ (unit vector) through the corneal apex and the pupil center, the lighting tone $\tau$ of the recording, and the subject $s$. Each training sample of a subject $s\in\{1,\dots,N_s\}$ provides multi-view images, an OCT volume and a labeled point cloud, from which $\mathbf{p}$, $r$ and $\mathbf{d}$ are measured. A new subject $s_{\mathrm{new}}$ is instead given by one or a few context images $\mathcal{I}_{\mathrm{ctx}}$. Formally, we learn the conditional distribution $p_\theta(\hat{I}_v, \hat{O}, \hat{P} \mid \mathcal{C}, \mathcal{I}_{\mathrm{ctx}})$. The simulator therefore serves two goals: (1) simulating learned and new subjects under the desired conditions, and (2) one- or few-shot reconstruction and novel-view rendering of a new subject from its context images.

Our EyeDiT consists of two parts, as illustrated in Fig.~\ref{fig:architecture}: a multi-modal anatomy VAE (Sec.~\ref{sec:method_vae}) that learns a shared eye latent space, and an eye latent DiT (Sec.~\ref{sec:method_dit}) that generates latents in this space from $\mathcal{C}$ and $\mathcal{I}_{\mathrm{ctx}}$.

\subsection{Multi-Modal Anatomy VAE}
\label{sec:method_vae}

The goal of our VAE is to learn a mapping from the near-eye images $\{I_v\}$, the OCT volume $O$ and the labeled point cloud $P$ to a compact shared latent $z_{\mathrm{fused}}$ that fuses the information from all modalities, on which our latent DiT is trained~\cite{Rombach2022LDM}. Inspired by query-based multi-modal fusion~\cite{Jaegle2022PerceiverIO}, we concatenate the tokens of the per-modality latents $z_{\mathrm{img}}$, $z_{\mathrm{oct}}$ and $z_{\mathrm{pcd}}$, and fuse them into $z_\text{fused}$ with a flexible attention-based fusion layer (Fig.~\ref{fig:architecture}).

\noindent \textbf{OCT Encoder.} An OCT volume $O \in \mathbb{R}^{D_o \times H_o \times W_o}$ is a stack of $D_o$ B-scans, each of which is a cross-sectional image of size $H_o \times W_o$ along one horizontal scan line. Since neighboring B-scans are strongly correlated in 3D, we encode $O$ with strided 3D convolution blocks that downsample it to the resolution of the latent grid, giving $z_{\mathrm{oct}} \in \mathbb{R}^{G \times d}$. Before the last block, the features are sampled at the scan position of each latent cell, given by the OCT device calibration. Here, $G$ denotes the number of cells in the latent grid, and $d$ is the token dimension.

\noindent \textbf{PCD Encoder.} Our point cloud is a set of labeled anatomy surface points, so we encode each label separately into its own $K$ latent tokens, using the same number of points per label and adding a learned label embedding to its points. Within each label, every point is embedded by a shared MLP~\cite{Qi2017PointNet} and pooled into the tokens with learnable queries and multi-head attention~\cite{Lee2019SetTransformer}, which keeps the encoding invariant to the point order. This gives $z_{\mathrm{pcd}} \in \mathbb{R}^{4K \times d}$, one group of tokens per anatomical surface.

\noindent \textbf{Image Encoder.} Our key design consideration for image encoding is to make it independent of the specific camera views and the number of views, which vary across datasets. Following recent multi-view transformers~\cite{Zhang2024GSLRM,Jin2025LVSM}, we generalize over views by adding to each image patch token the Pl\"ucker embedding of its camera rays~\cite{Sitzmann2021LFN}, computed from the calibration. The ray embedding makes each token 3D-aware, so that views are distinguished by their 3D geometry rather than by a view index. Given $N_v$ views, each split into $T$ patches, we obtain $z_{\mathrm{img}} \in \mathbb{R}^{N_v T \times d}$.

\noindent \textbf{Modality Fusion.} We obtain the 3D-layout latent $z_{\mathrm{fused}}$ with our latent fusion transformer. Given the concatenated per-modality tokens $[z_{\mathrm{img}}; z_{\mathrm{oct}}; z_{\mathrm{pcd}}] \in \mathbb{R}^{(N_v T + G + 4K) \times d}$, the transformer cross-attends to them with learnable 3D latent queries, one per cell of the latent grid, followed by self-attention among the queries. The fused latent $z_{\mathrm{fused}} \in \mathbb{R}^{G \times d}$ is thus arranged on a 3D grid of $G$ cells covering the eye region, and flexibly combines the information from all modalities. As in a standard VAE~\cite{Kingma2014VAE}, $z_{\mathrm{fused}}$ is sampled from a predicted Gaussian and regularized by a Kullback--Leibler (KL) divergence.

\noindent \textbf{Dual-Path Image Decoding.} To allow stable training of the 3DGS head for novel-view rendering, we decode images along two paths from the same latent: auxiliary camera-specific 2D image decoders and a 3DGS decoder. As suggested by~\cite{Charatan2024pixelSplat,Chen2024MVSplat,Zhang2024GSLRM,Ren2026TokenGS}, a typical approach for feed-forward novel-view rendering is to produce 3D Gaussians~\cite{Kerbl2023GS} while training them with images. However, due to the limited number of views in our training data, our method performs worse when trained with only the 3DGS head, as seen in Tab.~\ref{tab:abl_components}. We found that the shared latent, and thus the 3DGS head, can be trained in a more stable manner if we train the camera-specific 2D decoders and the 3DGS head together. For each camera-specific 2D decoder, we learn 2D grid queries, which implicitly learn to project the 3D eye tokens onto a 2D token grid of the corresponding camera. The 2D token grid is then decoded into the image $\hat{I}_v$ with 2D convolution layers. For the Gaussians, inspired by TokenGS~\cite{Ren2026TokenGS}, we introduce 3D Gaussian queries that are independent of a specific pixel layout. From the extracted Gaussian tokens, MLP layers produce Gaussians $\{G_k\}$, from which novel views can be rendered with any standard Gaussian rasterizer.

\noindent \textbf{OCT and Anatomy PCD Decoders.} In our setting, both OCT and PCD capture the anterior eye anatomy. Considering the dense 3D structure of OCT, we decode $z_{\mathrm{fused}}$ into 3D features with 3D up-convolution layers, which are then sampled at every OCT scan voxel using the OCT device calibration to produce $\hat{O}$. On the other hand, to explicitly retrieve each anatomical surface, we initialize learnable anatomy queries for each label and obtain anatomy tokens by cross-attending to the decoded 3D features. An MLP-based point cloud decoder then decodes the anatomy tokens of each label into its points, giving $\hat{P}$. The biometrics are then read from the decoded point cloud by fitting sphere and plane surfaces with RANSAC~\cite{Fischler1981RANSAC}.

\begin{table*}[!tp]
    \centering
    \caption{Reconstruction and generation quality on test-unseen subjects of \dsTwoV{} and \dsFourV{}. Values in parentheses are image metrics of the Gaussian render, and image metrics are averaged over all views. The conditioned DiT generates samples close to the VAE reconstructions in quality, suggesting that little fidelity is lost by generating from the conditions.}
    \label{tab:multi_modal_recon}
    \fitbox{0.87\textwidth}{%
        \begin{tabular}{llccccccccccc}
        \toprule
         & & \multicolumn{2}{c}{Image} & \multicolumn{2}{c}{OCT} & \multicolumn{2}{c}{PCD} & \multicolumn{5}{c}{Biometrics} \\
        \cmidrule(lr){3-4} \cmidrule(lr){5-6} \cmidrule(lr){7-8} \cmidrule(lr){9-13}
         & Dataset & PSNR (dB) $\uparrow$ & LPIPS $\downarrow$ & PSNR (dB) $\uparrow$ & LPIPS $\downarrow$ & Acc. (mm) $\downarrow$ & Comp. (mm) $\downarrow$ & Pos. (mm) $\downarrow$ & Rad. (mm) $\downarrow$ & OA ($^\circ$) $\downarrow$ & Cornea front $R$ (mm) $\downarrow$ & Iris thick. (mm) $\downarrow$ \\
        \midrule
         Recon. & \dsTwoV{} & 25.82 (25.42) & 0.101 (0.144) & 25.08 & 0.191 & 0.081 & 0.176 & 0.102 & 0.069 & 0.93 & 0.118 & 0.039 \\
        (VAE) & \dsFourV{} & 24.72 (23.62) & 0.135 (0.172) & 24.40 & 0.178 & 0.102 & 0.198 & 0.139 & 0.098 & 1.16 & 0.182 & 0.030 \\
        \midrule
        Gen. & \dsTwoV{} & 25.40 (24.76) & 0.106 (0.154) & 21.64 & 0.247 & 0.223 & 0.259 & 0.139 & 0.139 & 1.11 & 0.148 & 0.040 \\
        (DiT) & \dsFourV{} & 23.80 (22.57) & 0.143 (0.187) & 19.64 & 0.251 & 0.272 & 0.321 & 0.132 & 0.138 & 0.52 & 0.364 & 0.043 \\
        \bottomrule
        \end{tabular}%
    }%
\end{table*}

\subsection{Eye Latent DiT}
\label{sec:method_dit}

The goal of our eye latent DiT is to sample $z_{\mathrm{fused}}$ in the latent space learned by our VAE. Specifically, our DiT learns the conditional distribution $p_\theta(z_{\mathrm{fused}} \mid \mathcal{C}, \mathcal{I}_{\mathrm{ctx}})$ with the control conditions $\mathcal{C}$ and an optional set of context images.

\noindent \textbf{Condition Encoder.} We encode the pupil position $\mathbf{p}\in\mathbb{R}^3$, the optical axis $\mathbf{d}\in\mathbb{R}^3$, the pupil radius $r\in\mathbb{R}$ and the lighting tone $\tau\in\mathbb{R}$ with separate two-layer MLPs, and the subject identity $s$ with a learnable embedding $\mathbf{e}_s$ from a subject codebook. Each condition is embedded separately and summed with the timestep embedding $\phi_t(t)$ into a single condition embedding $z_{\mathrm{cond}}\in\mathbb{R}^{512}$,
\begin{equation}
\begin{aligned}
    z_{\mathrm{cond}} = {} & \phi_{\mathbf{p}}(\mathbf{p}) + \phi_{\mathbf{d}}(\mathbf{d}) + \phi_r(r) \\
    & + \phi_\tau(\tau) + \mathbf{e}_s + \phi_t(t)
\end{aligned}
\end{equation}
which conditions our DiT with adaLN-Zero~\cite{Peebles2023DiT}.

\noindent \textbf{Identity Representation.} Each training subject is represented by a learned embedding $\mathbf{e}_s\in\mathbb{R}^{512}$ in the subject codebook, with $s\in\{1,\dots,N_s\}$, which enters $z_{\mathrm{cond}}$. A new subject $s_{\mathrm{new}}$ is represented by one or more context images $\mathcal{I}_{\mathrm{ctx}}=\{I_v\}_{v\in\mathcal{S}_{\mathrm{ctx}}}$ from any view, while a learned null embedding $\mathbf{e}_\varnothing$ is assigned to its subject code. Since the identity is fixed by the subject code or the context latent, changing only the gaze or pupil conditions keeps the rest of the eye unchanged, which enables controlled simulation.

\noindent \textbf{Context-Guided Denoising.} The DiT combines the two identity paths, either the learned embedding $\mathbf{e}_s$ or the context images $\mathcal{I}_{\mathrm{ctx}}$, in each denoising step. Hence, we train our DiT to denoise with the added context latent $z_{\mathrm{ctx}}=\mathcal{E}_{\mathrm{ctx}}(\mathcal{I}_{\mathrm{ctx}})$, similar to input-level conditioning in latent diffusion~\cite{Rombach2022LDM,Zhang2023ControlNet}, jointly with the context encoder $\mathcal{E}_{\mathrm{ctx}}$. For the context encoder, we use the image encoder in our trained VAE as the initial checkpoint, and train it 
to predict a context latent $z_{\mathrm{ctx}}$ close to the target latent $z_{\mathrm{fused}}$,
\begin{equation}
    \mathcal{L}_{\mathrm{ctx}} = \| z_{\mathrm{ctx}} - z_{\mathrm{fused}} \|_2^2.
\end{equation}
Given the noisy latent $z_t = \alpha_t\, z_{\mathrm{fused}} + \sigma_t\, \epsilon$ at timestep $t$, with $\epsilon \sim \mathcal{N}(0, I)$, the DiT $f_\theta$ predicts the clean latent
\begin{equation}
\label{eq:denoise}
    \hat{z}_{\mathrm{fused}} = f_\theta\big(z_t + W_{\mathrm{ctx}}\, z_{\mathrm{ctx}};\ z_{\mathrm{cond}}\big),
\end{equation}
where $z_{\mathrm{ctx}}$ is projected by $W_{\mathrm{ctx}}$ and added to the input tokens, and $z_{\mathrm{cond}}$ modulates every block through adaLN-Zero~\cite{Peebles2023DiT}. The DiT thus generates the latent from the subject estimate $z_{\mathrm{ctx}}$ under the conditions $z_{\mathrm{cond}}$.

\begin{table}[!tb]
    \centering
    \caption{Image-modality comparison on \dsTwoV{} test-seen samples. Compared to baselines, our model reconstructs the given inputs better, and the distribution of our generated samples is closer to the real data.}
    \label{tab:image_modal_gen}
    \fitbox{\linewidth}{%
        \begin{tabular}{lcccccc}
        \toprule
         & \multicolumn{4}{c}{Reconstruction} & \multicolumn{2}{c}{Generation} \\
        \cmidrule(lr){2-5}
        \cmidrule(lr){6-7}
        Method & PSNR (dB) $\uparrow$ & LPIPS $\downarrow$ & rFD $\downarrow$ & rKD $\downarrow$ & gFD $\downarrow$ & gKD $\downarrow$ \\
        \midrule
        Lin et al.~\cite{Lin2026EyeTracker} & 17.36 & 0.440 & 255.48 & \textit{0.3325} & $\times$ & $\times$ \\
        GazePrior~\cite{Dumery2026GazePrior} & 17.30 & 0.442 & 362.64 & 0.5810 & 335.85 & 0.5270 \\
        ControllableGaze~\cite{Li2025ControllableGaze} & \textit{18.62} & \textit{0.354} & \textit{251.29} & 0.3658 & \textit{250.01} & \textit{0.3629} \\
        \textbf{Ours} & \textbf{26.30} & \textbf{0.077} & \textbf{67.69} & \textbf{0.0715} & \textbf{79.07} & \textbf{0.0917} \\
        \bottomrule
        \end{tabular}%
    }%
\end{table}

\subsection{Training Pipeline}
\label{sec:method_train}

Our training has two stages: we first train the VAE, and then the DiT and the context encoder on the latents of the frozen VAE. In the first stage, the VAE learns the latent space of $z_{\mathrm{fused}}$ by reconstructing all modalities $I_v$, $O$ and $P$ of each training sample jointly,
\begin{equation}
    \label{eq:vae_loss}
    \mathcal{L}_{\mathrm{VAE}} = \mathcal{L}_{\mathrm{oct}} + \mathcal{L}_{\mathrm{pcd}} + \mathcal{L}_{\mathrm{img}} + \mathcal{L}_{\mathrm{GS}} + \mathcal{L}_{\mathrm{KL}}.
\end{equation}
For OCT and images, $\mathcal{L}_{\mathrm{oct}}$ and $\mathcal{L}_{\mathrm{img}}$ combine L1 and LPIPS~\cite{Zhang2018LPIPS} losses between the decoded and the input images, where each B-scan slice of the OCT volume is treated as an image. For the point cloud, we use the symmetric Chamfer distance, $\mathcal{L}_{\mathrm{pcd}}=\mathrm{CD}(\hat{P},P)$. For the Gaussians, $\mathcal{L}_{\mathrm{GS}}$ applies the same L1 and LPIPS losses to the renders $\tilde{I}_v$ inside the field-of-view mask. $\mathcal{L}_{\mathrm{KL}}$ is the KL divergence of the posterior from a unit Gaussian.

In the second stage, the DiT predicts the clean latent from the noised latent $z_t$ of Eq.~\ref{eq:denoise}~\cite{Li2026JiT},
\begin{equation}
    \mathcal{L}_{\mathrm{DiT}} = \mathbb{E}_{t,\epsilon}\big[\, w_t\, \| \hat{z}_{\mathrm{fused}} - z_{\mathrm{fused}} \|_2^2 \big] + \mathcal{L}_{\mathrm{ctx}}.
\end{equation}
Here, $w_t$ weights the error as a velocity loss, and $\mathcal{L}_{\mathrm{ctx}}$ trains the context encoder (Sec.~\ref{sec:method_dit}).

\noindent \textbf{Condition Dropout.} To let one model generate from the conditions alone, from context images, or from both, we randomly replace the context images $\mathcal{I}_{\mathrm{ctx}}$, the subject embedding $\mathbf{e}_s$ and the pupil conditions $(\mathbf{p}, r, \mathbf{d})$ with learned null embeddings $\varnothing$ during training. Dropping the pupil conditions and the subject together also enables classifier-free guidance~\cite{Ho2022CFG} at sampling,
\begin{equation}
    \hat{z}^{w}_{\mathrm{fused}} = f_\theta(\cdot\,; z_{\mathrm{cond}}^{\varnothing}) + w\,\big(f_\theta(\cdot\,; z_{\mathrm{cond}}) - f_\theta(\cdot\,; z_{\mathrm{cond}}^{\varnothing})\big),
\end{equation}
applied to the predicted clean latent with $w=1.25$, where $\cdot$ is the DiT input of Eq.~\ref{eq:denoise} and $z_{\mathrm{cond}}^{\varnothing}$ uses the null embeddings for the pupil and subject conditions.

\section{Experiments}

\subsection{Datasets}

\noindent \textbf{Datasets} We use four datasets: \dsTwoV{}, \dsFourV{}, the light-dome capture of Lin et al.~\cite{Lin2026EyeTracker}, which we refer to as \dsMasq{}, and \dsNymeria{}~\cite{Ma2024Nymeria}. We collected \dsTwoV{} (two cameras, 78 subjects, 7,931 volumes) and \dsFourV{} (four cameras, 50 subjects, 8,265 volumes), in which near-eye cameras capture the eye together with an anterior-segment OCT volume while the subject fixates targets. Post-processing the OCT volumes yields metric 3D pupil positions, pupil radii, optical axes and segmented anatomy point clouds. \dsMasq{}~\cite{Lin2026EyeTracker} contains images from 18 close-up cameras, without OCT volumes. \dsTwoV{} and \dsFourV{} are split into 6,331 / 829 / 771 and 6,460 / 800 / 1,005 volumes for training, test-seen (held-out samples of training subjects) and test-unseen (new subjects), and we evaluate on fixed subsets of the test splits. The VAE is trained on the training splits of the first three datasets, and the DiT on those of \dsTwoV{} and \dsFourV{}, since \dsMasq{} has no pupil or optical-axis labels. For eye-tracking evaluation on a public device, we use the eye-tracking images of \dsNymeria{}~\cite{Ma2024Nymeria}, recorded with Project Aria glasses~\cite{Engel2023ProjectAria} (Sec.~\ref{sec:s2r_main}). Details are in the appendix.

\subsection{Multi-Modal Generation}

    Our model reconstructs and generates all modalities from one latent (Tab.~\ref{tab:multi_modal_recon}, Fig.~\ref{fig:multi_modal_recon}). Reconstruction (VAE) decodes the given inputs, and generation (DiT) produces the same samples from the conditions and one nasal context image in \dsTwoV{} and \dsFourV{}. The reconstructed OCT volumes and point clouds closely match the inputs, and the generated outputs stay close to the reconstruction, for example 25.4 against 25.8\,dB image PSNR and 0.14 against 0.10\,mm pupil-position error on \dsTwoV{}. Since our model generates the anatomy point cloud together with the images, the biometrics of each sample can be read directly from it. For example, the optical-axis errors are 1.1$^\circ$ and 0.5$^\circ$ on \dsTwoV{} and \dsFourV{}. As the DiT is conditioned on the optical axis, these errors measure how closely the generated anatomy follows the commanded axis.

\noindent \textbf{Baseline Comparison} As shown in Tab.~\ref{tab:image_modal_gen}, we compare the image modality of our model with previous synthetic eye image generators, Lin et al.~\cite{Lin2026EyeTracker}, GazePrior~\cite{Dumery2026GazePrior} and ControllableGaze~\cite{Li2025ControllableGaze}, on \dsTwoV{}. Our model reaches a much higher reconstruction PSNR (26.3 against at most 18.6\,dB), and its generated samples are much closer to the real data in distribution (Fr\'echet distance~\cite{Heusel2017FID,Stein2023Exposing} of 79.1 against at least 250.0). Unlike these generators, our model also generates the OCT volume and the anatomy point cloud of each sample together with its images.

\begin{figure}[!t]
    \centering
    \includegraphics[width=0.6\linewidth]{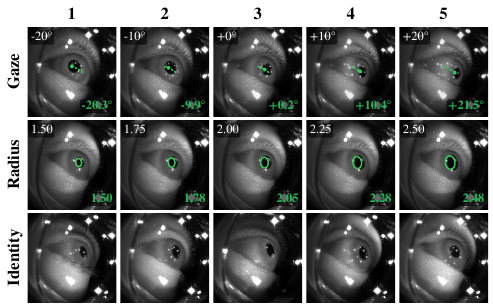}
    \caption{Generation results when changing the gaze, the pupil radius and the identity embedding, respectively (rows), in five steps (columns). Only the changed factor varies along each row, and the generated pupil follows the commanded values. The green lines and circles mark the commanded optical axis and pupil radius. Top-left: commanded value. Bottom-right: readout value.}
    \label{fig:qual_condition_sweep}
    \vspace{-15pt} 

\end{figure}

\begin{table*}[!t]
    \centering
    \caption{Novel-view rendering on \dsTwoV{}, \dsFourV{} and \dsMasq{}~\cite{Lin2026EyeTracker}. The top three methods are baselines for eye modeling, while the others are general 3D reconstruction and generation methods. Our model, with a prior learned from multi-view eye data, shows the best novel-view synthesis performance. $\times$: the method needs at least two context views. --: ControllableGaze is not available on \dsMasq{}, which has no mesh or OCT data.}
    \label{tab:nv_merged}
    \fitbox{0.87\textwidth}{%
    \begin{tabular}{llccccccccccccccc}
    \toprule
     & & \multicolumn{3}{c}{\dsTwoV{}} & \multicolumn{6}{c}{\dsFourV{}} & \multicolumn{6}{c}{\dsMasq{}} \\
    \cmidrule(lr){3-5}
    \cmidrule(lr){6-11}
    \cmidrule(lr){12-17}
     & & \multicolumn{3}{c}{1 view} & \multicolumn{3}{c}{1 view} & \multicolumn{3}{c}{3 views} & \multicolumn{3}{c}{1 view} & \multicolumn{3}{c}{3 views} \\
    \cmidrule(lr){3-5}
    \cmidrule(lr){6-8}
    \cmidrule(lr){9-11}
    \cmidrule(lr){12-14}
    \cmidrule(lr){15-17}
    Method & Source & PSNR $\uparrow$ & SSIM $\uparrow$ & LPIPS $\downarrow$ & PSNR $\uparrow$ & SSIM $\uparrow$ & LPIPS $\downarrow$ & PSNR $\uparrow$ & SSIM $\uparrow$ & LPIPS $\downarrow$ & PSNR $\uparrow$ & SSIM $\uparrow$ & LPIPS $\downarrow$ & PSNR $\uparrow$ & SSIM $\uparrow$ & LPIPS $\downarrow$ \\
    \midrule
    GazePrior~\cite{Dumery2026GazePrior} & OCT-Cam & 20.23 & 0.694 & 0.468 & \textit{27.13} & \textit{0.784} & 0.502 & \textit{27.54} & \textit{0.788} & 0.499 & \textit{16.08} & \textit{0.485} & \textit{0.576} & \textit{17.27} & 0.493 & \textit{0.554} \\
    Lin et al.~\cite{Lin2026EyeTracker} & Static Fit & 14.15 & 0.267 & 0.595 & 11.77 & 0.375 & 0.665 & 15.85 & 0.480 & 0.582 & 11.73 & 0.404 & 0.723 & 14.22 & 0.405 & 0.588 \\
    ControllableGaze~\cite{Li2025ControllableGaze} & OCT-Cam & \textit{21.81} & \textit{0.702} & 0.397 & 19.24 & 0.726 & \textit{0.463} & 18.93 & 0.725 & \textit{0.466} & -- & -- & -- & -- & -- & -- \\
    \midrule
     MapAnything~\cite{Keetha2026MapAnything} (+GS Head) & Pretrained (+OCT-Cam) & 19.62 & 0.653 & \textit{0.355} & 13.66 & 0.575 & 0.500 & 14.50 & 0.681 & 0.481 & 10.73 & 0.353 & 0.806 & 12.32 & 0.451 & 0.758 \\
    MoGe-2~\cite{Wang2025MoGe2} (+GS Head) & Pretrained (+OCT-Cam) & 19.10 & 0.630 & 0.395 & 15.35 & 0.576 & 0.516 & 17.95 & 0.685 & 0.468 & 14.83 & 0.439 & 0.581 & 10.33 & 0.398 & 0.790 \\
    DA-V2~\cite{Yang2024DepthAnythingV2} (+GS Head) & Pretrained (+OCT-Cam) & 18.82 & 0.634 & 0.439 & 14.08 & 0.350 & 0.677 & 15.10 & 0.488 & 0.665 & 10.23 & 0.267 & 0.795 & 12.38 & 0.413 & 0.682 \\
    MVSplat~\cite{Chen2024MVSplat} & Pretrained & $\times$ & $\times$ & $\times$ & $\times$ & $\times$ & $\times$ & 14.34 & 0.490 & 0.621 & $\times$ & $\times$ & $\times$ & 10.68 & 0.102 & 0.746 \\
    DepthSplat~\cite{Xu2025DepthSplat} & Pretrained & $\times$ & $\times$ & $\times$ & $\times$ & $\times$ & $\times$ & 14.56 & 0.406 & 0.664 & $\times$ & $\times$ & $\times$ & 12.15 & 0.417 & 0.767 \\
    TRELLIS~\cite{Xiang2025TRELLIS} & Pretrained & 15.20 & 0.514 & 0.581 & 13.38 & 0.229 & 0.653 & 15.74 & 0.525 & 0.555 & 9.68 & 0.309 & 0.826 & 15.07 & \textbf{0.521} & 0.718 \\
    ViewCrafter~\cite{Yu2025ViewCrafter} & Pretrained & 15.56 & 0.501 & 0.514 & 8.07 & 0.439 & 0.711 & 13.14 & 0.514 & 0.676 & 11.95 & 0.298 & 0.658 & 10.90 & 0.323 & 0.775 \\
    \midrule
     \textbf{Ours} & OCT-Cam & \textbf{24.89} & \textbf{0.786} & \textbf{0.128} & \textbf{29.23} & \textbf{0.891} & \textbf{0.137} & \textbf{29.42} & \textbf{0.894} & \textbf{0.130} & \textbf{18.08} & \textbf{0.502} & \textbf{0.358} & \textbf{19.30} & \textit{0.520} & \textbf{0.324} \\
    \bottomrule
    \end{tabular}%
    }%
    \vspace{4pt}
    \centering
        \includegraphics[width=\textwidth]{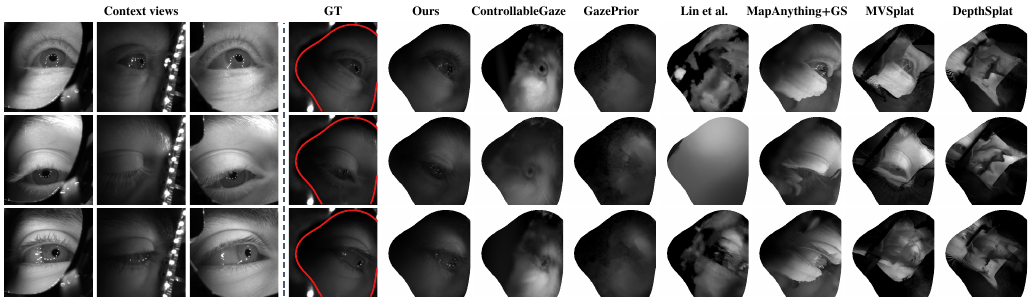}
    \captionof{figure}{Novel-view rendering of our method and the baselines on \dsFourV{}. Three context views (left of the dashed line) are used as inputs to render the target view (GT). Pixels outside the red lines in the GT column are occluded by our capture device and excluded from novel-view evaluation. Our method renders high-fidelity novel views even from only three input views, while ControllableGaze~\cite{Li2025ControllableGaze} and GazePrior~\cite{Dumery2026GazePrior} show significantly blurrier results in this few-shot setting.}

    \label{fig:qual_nv_main}
\end{table*}

\noindent \textbf{Condition Sweeps} As seen in Fig.~\ref{fig:qual_condition_sweep}, each condition controls its own factor of generation: varying the gaze (optical axis) or the pupil radius changes only that factor, following the commanded value, while the rest of the eye is preserved, and varying the identity embedding changes the subject.

\subsection{Novel-View Synthesis}

\noindent \textbf{Baselines} We compare with eye image synthesis methods and general novel-view synthesis methods (Tab.~\ref{tab:nv_merged}). For the eye image synthesis domain, we use Lin et al.~\cite{Lin2026EyeTracker}, which fits an eye mesh and a neural radiance field to each capture, and GazePrior~\cite{Dumery2026GazePrior} and ControllableGaze~\cite{Li2025ControllableGaze}, which we train on the training splits of \dsTwoV{} and \dsFourV{} as for our model. For general methods, we use the feed-forward Gaussian methods MVSplat~\cite{Chen2024MVSplat} and DepthSplat~\cite{Xu2025DepthSplat}, the geometry foundation models~\cite{Keetha2026MapAnything, Wang2025MoGe2, Yang2024DepthAnythingV2}, and the generative methods~\cite{Xiang2025TRELLIS, Yu2025ViewCrafter}. The geometry foundation models have no Gaussian head, so we train one for each of them on our training splits. As target views, we use the temporal view on \dsTwoV{} and \dsFourV{}, and camera 9 on \dsMasq{}.

\noindent \textbf{Results} Tab.~\ref{tab:nv_merged} shows that our model achieves the highest PSNR among the compared methods, even on \dsMasq{}, on which our DiT is not trained due to the unavailability of OCT and biometric annotations. The general pretrained methods all stay below 20\,dB PSNR and often miss the eye structure (Fig.~\ref{fig:qual_nv_main}), and Lin et al., which optimizes each scene separately, reaches at most 15.9\,dB in this few-view setting. Thanks to the prior learned from our training splits, GazePrior~\cite{Dumery2026GazePrior} achieves a fairly high reconstruction quality on the OCT-Cam datasets, reaching 20.2 and 27.1\,dB with one context view. Compared to GazePrior, our model with feed-forward prediction from a learned prior achieves an even higher PSNR of 24.9 and 29.2\,dB.

\subsection{Eye Tracking Training from Synthetic Data}
\label{sec:s2r_main}

A direct use of a synthetic eye generator is training eye-tracking (ET) models such as the Aria ET estimator~\cite{Engel2023ProjectAria,Mansour2025AriaET}, which require large and diverse labeled data. We evaluate this use on \dsTwoV{} and \dsNymeria{}~\cite{Ma2024Nymeria}.

\begin{figure}[!t]
    \centering
    \includegraphics[width=0.6\linewidth]{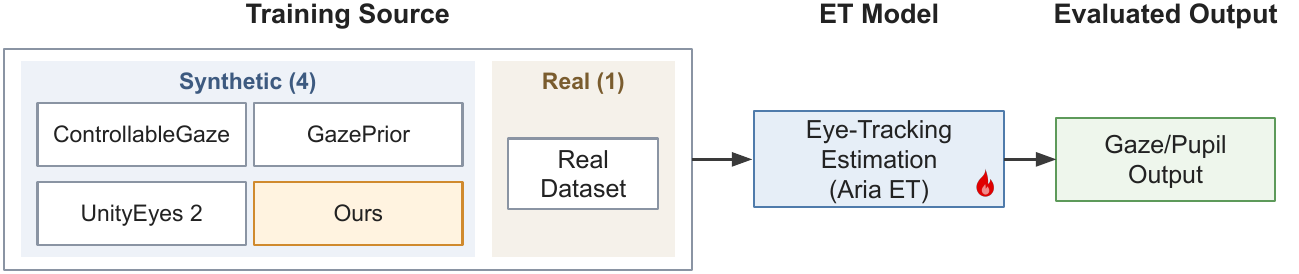}
    \caption{ET training protocol. An Aria ET model is trained on synthetic data from different sources, or on real data, and tested on real data to measure the sim-to-real gap.}
    \label{fig:sim_to_real_overview}
\end{figure}

\noindent \textbf{Protocol} Each recording of \dsTwoV{} is split by time into source, train and test parts, with 2,936, 2,817 and 578 volumes. Our model and each learned baseline~\cite{Li2025ControllableGaze,Dumery2026GazePrior} are trained on the source part and synthesize the train part from the conditions of its real frames. For UnityEyes 2~\cite{Smith2025UnityEyes2}, we give these conditions to its parametric eye model. An ET model is then trained on each synthetic set, or on the real train part as an oracle, and evaluated on the real test part (Fig.~\ref{fig:sim_to_real_overview}). On \dsNymeria{}, we use 7,102 training and 158 test frames with gaze labels, and synthesize each of two folds of training participants with a generator fine-tuned on the other fold. 
All training sources use the Aria ET architecture~\cite{Mansour2025AriaET}, extended with 3D gaze and pupil heads on \dsTwoV{}.

\begin{table}[!t]
    \centering
    \caption{Aria ET~\protect\cite{Mansour2025AriaET} errors on real \dsTwoV{} and \dsNymeria{}~\protect\cite{Ma2024Nymeria} images, by training source. The estimator trained on our synthetic data has the smallest gap to the one trained on real data. \dsNymeria{} uses gaze (visual-axis) labels. $\times$: ControllableGaze is not available on \dsNymeria{}, which lacks mesh and OCT data.}
    \label{tab:sim_to_real_aria}
    \fitbox{0.87\linewidth}{%
        \begin{tabular}{lcccccccccccc}
        \toprule
         & \multicolumn{9}{c}{\dsTwoV{}} & \multicolumn{3}{c}{\dsNymeria{}} \\
        \cmidrule(lr){2-10} \cmidrule(lr){11-13}
         & \multicolumn{3}{c}{Pupil pos. (mm) $\downarrow$} & \multicolumn{3}{c}{Pupil rad. (mm) $\downarrow$} & \multicolumn{3}{c}{Optical axis ($^\circ$) $\downarrow$} & \multicolumn{3}{c}{Gaze ($^\circ$) $\downarrow$} \\
        \cmidrule(lr){2-4} \cmidrule(lr){5-7} \cmidrule(lr){8-10} \cmidrule(lr){11-13}
        ET trained on & E50 & E75 & E95 & E50 & E75 & E95 & E50 & E75 & E95 & E50 & E75 & E95 \\
        \midrule
        ControllableGaze~\cite{Li2025ControllableGaze} & \textit{3.20} & \textit{4.32} & \textit{6.44} & 0.65 & 1.02 & 1.49 & \textit{9.36} & \textit{12.70} & \textit{18.70} & $\times$ & $\times$ & $\times$ \\
        GazePrior~\cite{Dumery2026GazePrior} & 18.96 & 21.17 & 24.46 & 0.58 & 0.95 & 1.53 & 16.68 & 23.69 & 36.64 & \textit{8.41} & \textit{12.05} & \textit{18.89} \\
        UnityEyes 2~\cite{Smith2025UnityEyes2} & 10.55 & 12.37 & 16.69 & \textit{0.57} & \textit{0.90} & \textit{1.42} & 12.90 & 18.83 & 27.40 & 10.18 & 13.89 & 24.00 \\
        \textbf{Ours} & \textbf{0.61} & \textbf{0.85} & \textbf{1.33} & \textbf{0.34} & \textbf{0.43} & \textbf{0.59} & \textbf{4.17} & \textbf{6.01} & \textbf{9.16} & \textbf{3.53} & \textbf{6.18} & \textbf{14.91} \\
        \midrule
        Real (oracle) & 0.64 & 0.80 & 1.12 & 0.25 & 0.31 & 0.41 & 2.84 & \textbf{4.21} & \textbf{6.20} & 2.59 & 4.17 & 8.50 \\
        Real + Ours & \textbf{0.44} & \textbf{0.58} & \textbf{0.90} & \textbf{0.10} & \textbf{0.15} & \textbf{0.25} & \textbf{2.75} & 4.29 & 6.91 & \textbf{2.28} & \textbf{3.48} & \textbf{7.47} \\
        \bottomrule
        \end{tabular}%
    }%
\end{table}

\noindent \textbf{Results} On \dsTwoV{} (Tab.~\ref{tab:sim_to_real_aria}), the estimator trained on our data has the lowest pupil-position, pupil-radius and optical-axis errors among the synthetic sources and comes closest to the real-data oracle. GazePrior and UnityEyes 2 give large pupil-position errors, as GazePrior renders blurry images in which the pupil boundary is barely visible and UnityEyes 2 differs strongly from the real IR images in appearance. On \dsNymeria{}, our data also gives lower gaze errors than the other synthetic sources, with an E50 of $3.53^\circ$ against $2.59^\circ$ for the oracle. Adding our data to the real training images lowers the pupil-position E50 from 0.64 to 0.44\,mm on \dsTwoV{} and the gaze E50 from $2.59^\circ$ to $2.28^\circ$ on \dsNymeria{}. Although the generator never sees the test participants of \dsNymeria{}, our data still improves the real-data ET model.

\subsection{Ablation Study}
\label{sec:ablation}

\noindent \textbf{Modality Ablation} Tab.~\ref{tab:abl_modality} ablates the anatomy modalities. When the PCD is not generated, reading the biometrics (pupil position, pupil radius and optical axis) is not straightforward. When only images are available, we instead use multi-view triangulation to estimate the pupil position and radius. When OCT is available, we apply the same OCT annotation engine to the generated OCT B-scans to obtain the biometrics. The image-based biometric estimation is much less accurate than any setting with PCD or OCT. In particular, our method with both OCT and PCD results in the smallest errors, thanks to the accurate biometric reading from the point cloud outputs. Meanwhile, the image quality decreases only slightly with these additional modalities (24.76 vs.\ 25.27\,dB). Without the anatomy modalities, the biometrics can only be estimated from the generated images, which makes them far less accurate or unavailable.

\noindent \textbf{Component Ablation} Tab.~\ref{tab:abl_components} ablates our model components on unseen \dsTwoV{} subjects. Without OCT device warping, the OCT encoder and decoder cannot align scans from different devices and recordings with the latent grid, which severely degrades the OCT LPIPS (0.247 to 0.435) and the optical axis (1.11$^\circ$ to 3.33$^\circ$). Without the auxiliary 2D decoders, both the image quality and most biometrics degrade, suggesting that training the 3DGS head alone hurts the 3D consistency of the latent. Finally, without condition dropout, i.e., when all conditions and the context image are always given during training, the model never learns to generate from the conditions alone, so the biometrics generated without context severely degrade (5.12$^\circ$ optical-axis error).

\begin{table}[!t]
    \centering
    \caption{Ablation of the model components on test-unseen \dsTwoV{} subjects. Base denotes our full model.}
    \label{tab:abl_components}
    \fitbox{\linewidth}{%
    \begin{tabular}{lccccc}
    \toprule
    Variant & Img PSNR $\uparrow$ & OCT LPIPS $\downarrow$ & PCD Acc. $\downarrow$ & Rad. (mm) $\downarrow$ & OA ($^\circ$) $\downarrow$ \\
    \midrule
    Base & \textbf{24.76} & \textbf{0.247} & 0.223 & 0.139 & \textbf{1.11} \\
    w/o OCT device warping & 24.34 & 0.435 & 0.292 & \textbf{0.124} & 3.33 \\
    w/o 2D decoders & 22.27 & 0.489 & \textbf{0.199} & 0.269 & 1.45 \\
    w/o condition dropout & 21.34 & 0.376 & 0.470 & 0.412 & 5.12 \\
    \bottomrule
    \end{tabular}%
    }
\end{table}

\begin{table}[!t]
    \centering
    \caption{Training-modality ablation on test-unseen \dsTwoV{} subjects, generated from one context image (N). Img PSNR: Gaussian render in the FoV mask, averaged over both views as in Tab.~\ref{tab:multi_modal_recon}. Image + OCT: anatomy read from the generated OCT. Image: anatomy read from the generated images.}
    \label{tab:abl_modality}
    \fitbox{\linewidth}{%
    \begin{tabular}{lccccc}
    \toprule
    Training data & Img PSNR $\uparrow$ & OCT LPIPS $\downarrow$ & Pos. (mm) $\downarrow$ & Rad. (mm) $\downarrow$ & OA ($^\circ$) $\downarrow$ \\
    \midrule
    Image & 25.27 & -- & 1.206 & 0.226 & -- \\
    Image + PCD & \textbf{25.37} & -- & 0.146 & 0.191 & 1.21 \\
    Image + OCT & 25.31 & 0.253 & 0.256 & 0.276 & 3.15 \\
    Image + OCT + PCD (Base) & 24.76 & \textbf{0.247} & \textbf{0.139} & \textbf{0.139} & \textbf{1.11} \\
    \bottomrule
    \end{tabular}%
    }
\end{table}

\section{Conclusion}

    We presented EyeDiT, a multi-modal eye model that generates synthetic near-eye images grounded in anatomical priors from OCT and labeled point clouds. Unlike prior generators that mainly control gaze, it controls the gaze, the pupil radius, the subject identity and the viewpoint, and provides rich anatomical labels. A public ET model~\cite{Mansour2025AriaET} trained on our data approaches training on real data, and augmenting the real data with ours further improves it, which could reduce the data collection cost for new AR/VR hardware.

\noindent \textbf{Limitations and Future Work} Our model does not learn temporal eye movement patterns, but simulates the eye in a quasi-static manner. Dynamic modeling of the eye may further improve the efficiency and realism. Separate modeling of the eyeball and the periocular region (eyelids and skin) is also left for future work, as it could better isolate gaze and pupil changes, which mostly affect the eyeball region.

\clearpage
\newpage
\section*{Acknowledgments}

We are grateful to our former colleagues, in particular Mohamed El-Haddad, Francesco La Rocca, Ruobing Qian, Catherine Fromm, Yimin Ding and Weihan Zhang, for their legacy work on the OCT setup and the OCT engine for eye tracking.

\bibliographystyle{assets/plainnat}
\bibliography{paper}

\clearpage
\newpage
\beginappendix

\newcommand{\suppfig}[2]{%
    \begin{figure}[!ht]
        \centering
        \includegraphics[width=0.85\textwidth,height=0.82\textheight,keepaspectratio]{figures/compressed/#1_compressed.pdf}
        \caption{#2}
        \label{fig:#1}
    \end{figure}}

\section{Dataset Details}
\label{sec:supp_datasets}
\noindent \textbf{Dataset Collection} \dsTwoV{} and \dsFourV{} were captured with our OCT-paired camera rigs. While the subject fixates targets, near-eye cameras record the eye at the same time as an anterior-segment OCT volume, with two cameras in \dsTwoV{} and four in \dsFourV{}. We denote the \dsTwoV{} cameras as nasal (N) and temporal (T), and the \dsFourV{} cameras as stereo nasal (SN), stereo temporal (ST), direct temporal (DT) and an additional temporal camera (AT). \dsTwoV{} has 78 subjects and 653 recordings (7{,}931 volumes), and \dsFourV{} has 50 subjects and 437 recordings (8{,}265 volumes). An OCT post-processing engine lifts each volume to an anatomy point cloud and its biometrics, including the pupil position, pupil radius and optical axis. \dsMasq{}~\cite{Lin2026EyeTracker} and \dsNymeria{}~\cite{Ma2024Nymeria} are existing image datasets without OCT.

\noindent \textbf{Devices and Calibration} The OCT scans the anterior segment in a bidirectional raster pattern with galvanometer mirrors. Each volume has 100 B-scans of 300 A-scans with 3328 depth samples (about 0.015\,mm per sample). We crop each volume in depth to a window around the cornea and the iris and resize it to $100\times64\times128$ voxels, keeping all 100 B-scans. The near-eye cameras capture $400\times400$ images in \dsTwoV{}, and $800\times800$ (stereo pair, cropped from $1280\times800$) and $384\times384$ images in \dsFourV{}. All images are resized to $256\times256$. The OCT and the cameras are calibrated with a dot-grid target imaged by both at several poses, and a rigid OCT-to-camera transform is fitted on the matched dots, with a held-out alignment error of 0.07\,mm on \dsTwoV{}.

\noindent \textbf{Anatomy Annotation} Following Qian et al.~\cite{Qian2026WEOCT}, each OCT volume is lifted to a 3D point cloud, and its points are annotated with five labels: the front and back surfaces of the cornea, the front and back surfaces of the iris, and the sclera. We use the four cornea and iris labels and leave out the sclera. Fig.~\ref{fig:s1_annotation} shows an example of the raw near-eye images and OCT data, the lifted point cloud with its labels, the pupil annotation and the optical axes.

\newlength{\annh}
\setlength{\annh}{0.155\textwidth}
\begin{figure}[!ht]
    \centering
    \begin{minipage}[b]{0.153\textwidth}\centering
        \includegraphics[width=\linewidth]{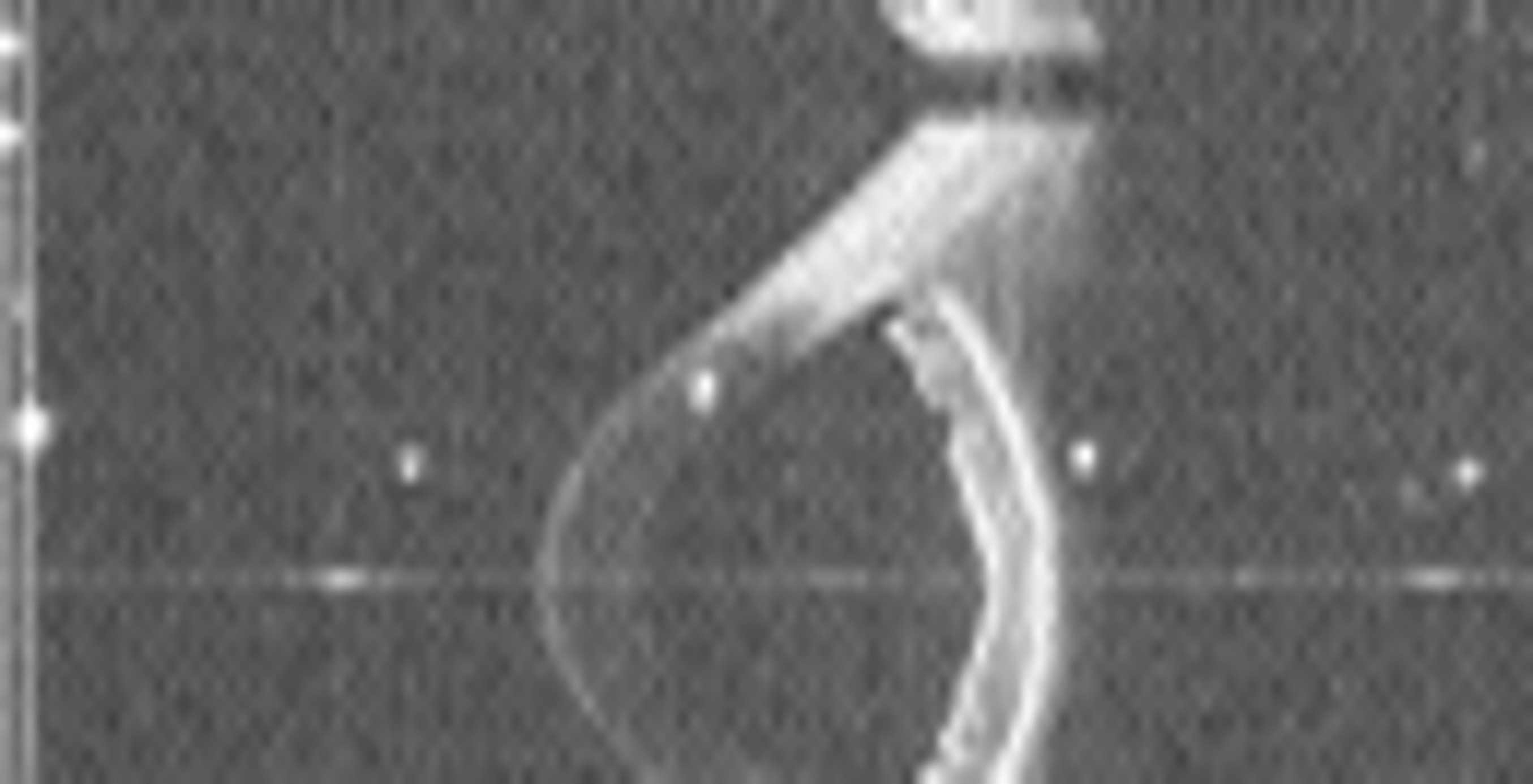}\\[2pt]
        \includegraphics[width=\linewidth]{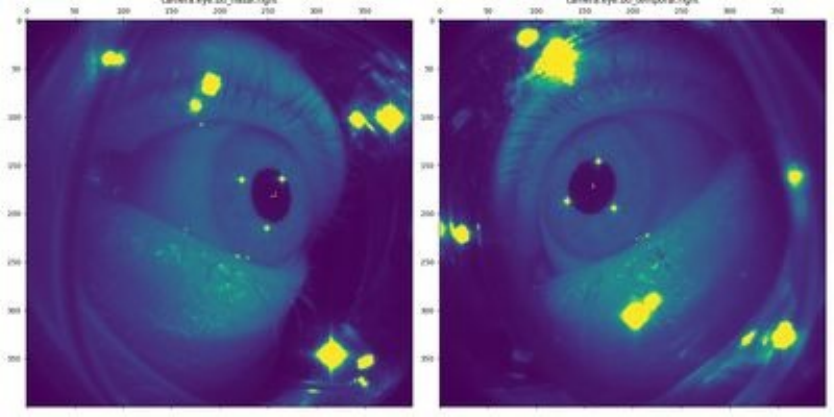}\\[2pt]
        {\small (a) Input Data}
    \end{minipage}\hfill
    \begin{minipage}[b]{0.19\textwidth}\centering
        \includegraphics[height=\annh]{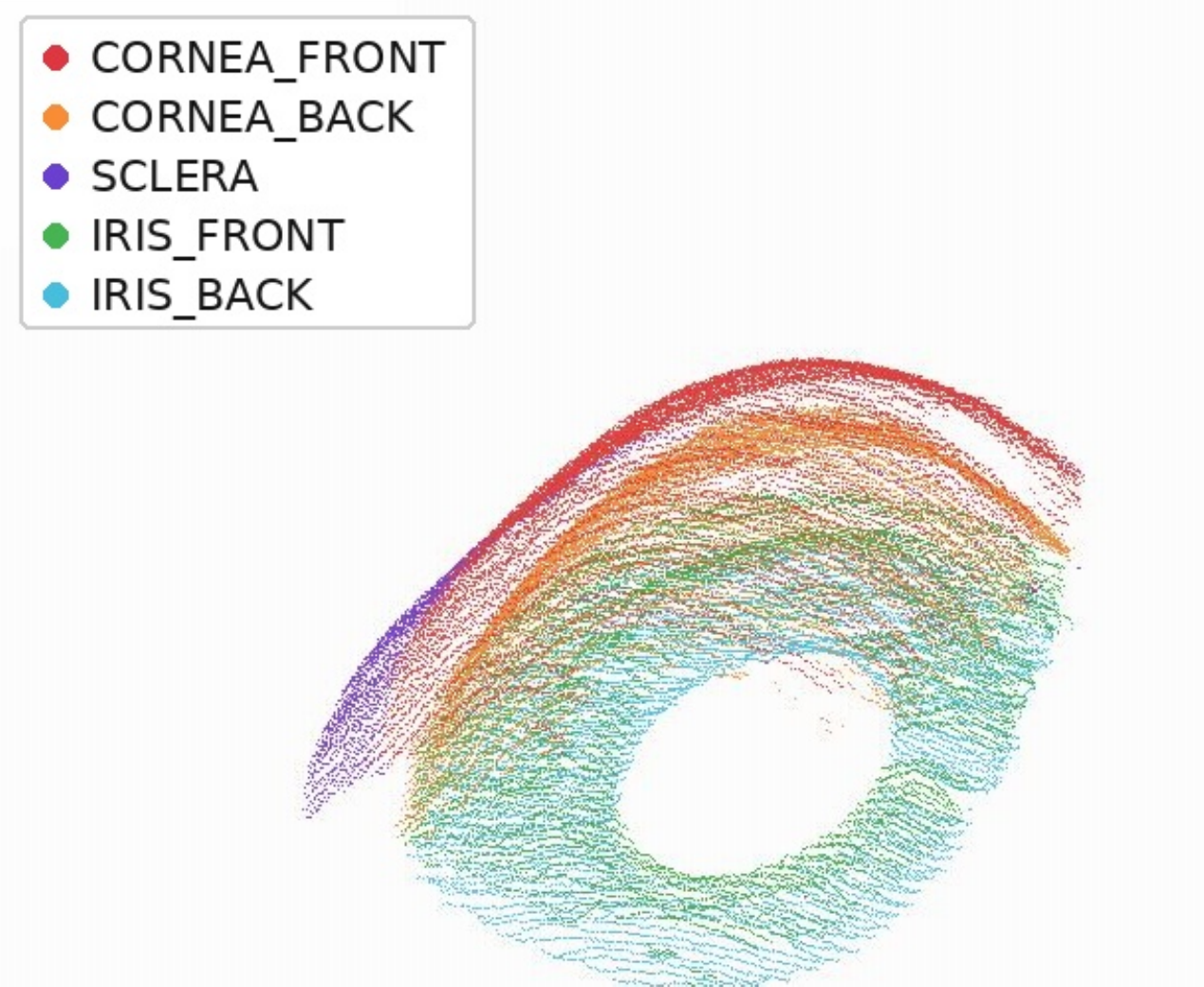}\\[2pt]
        {\small (b) Lifted OCT PCD}
    \end{minipage}\hfill
    \begin{minipage}[b]{0.37\textwidth}\centering
        \includegraphics[height=\annh]{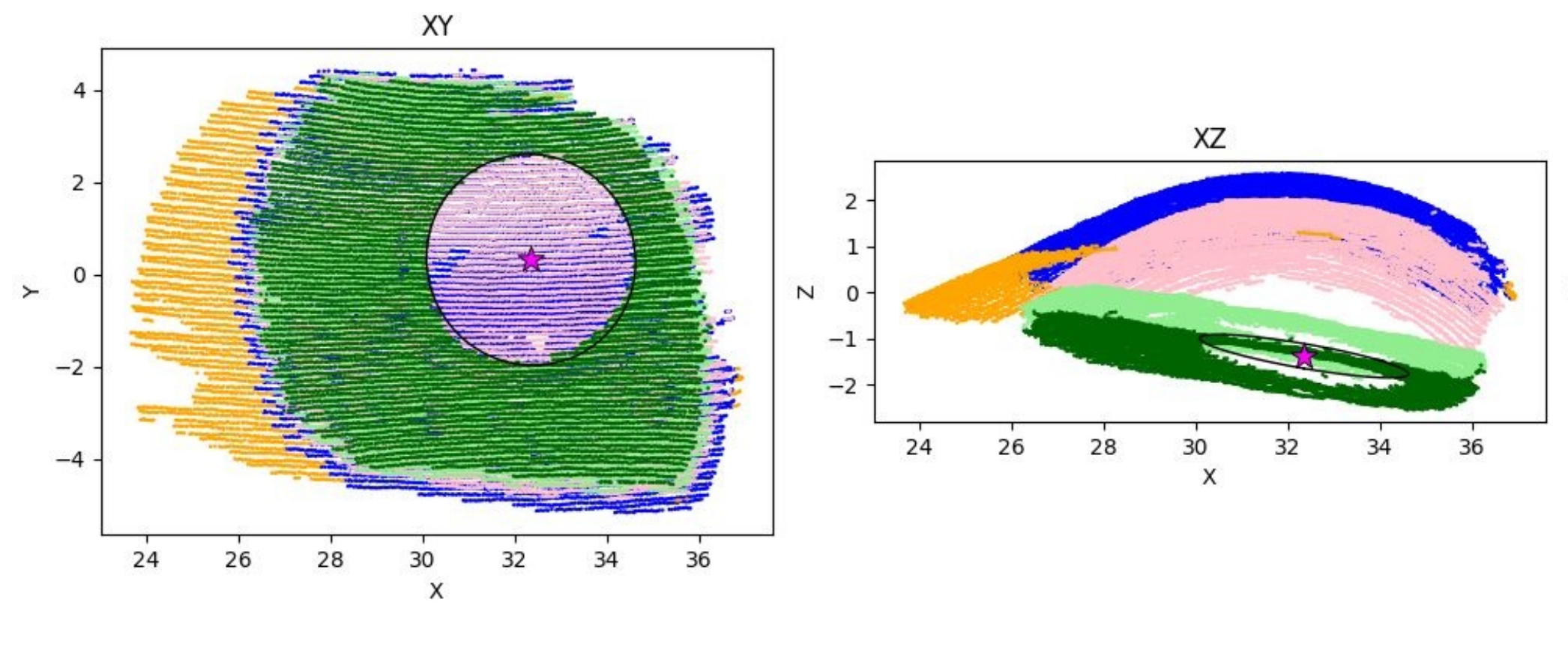}\\[2pt]
        {\small (c) Pupil Annotation}
    \end{minipage}\hfill
    \begin{minipage}[b]{0.175\textwidth}\centering
        \includegraphics[height=\annh]{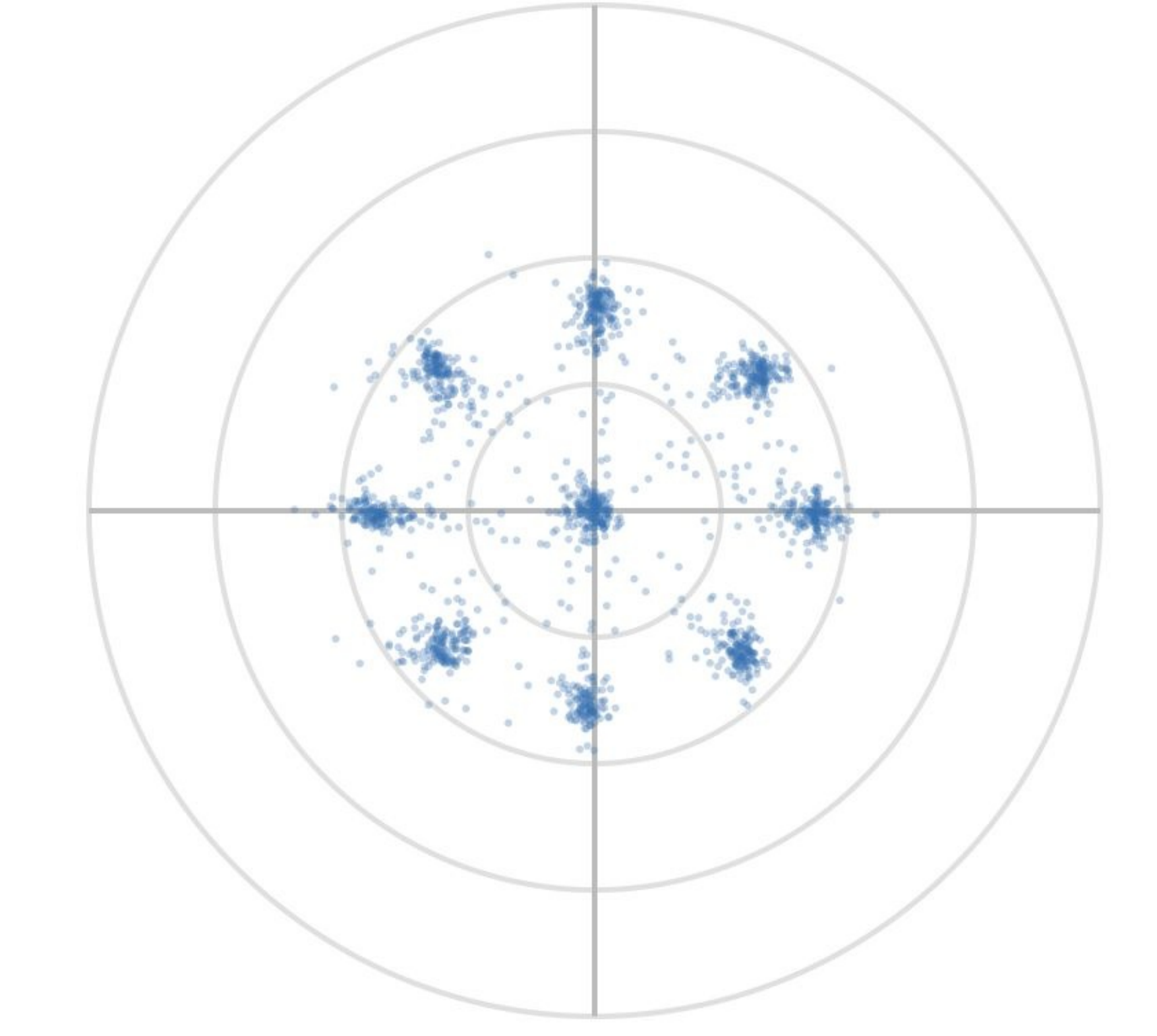}\\[2pt]
        {\small (d) Optical Axis}
    \end{minipage}
    \caption{Raw data and anatomy annotation. (a) A raw OCT B-scan and the near-eye camera images. (b) The point cloud lifted from the OCT volume with its five labels. (c) The pupil annotation on the point cloud. (d) The optical axes.}
    \label{fig:s1_annotation}
\end{figure}

\begin{figure}[!ht]
    \centering
    \includegraphics[width=\textwidth]{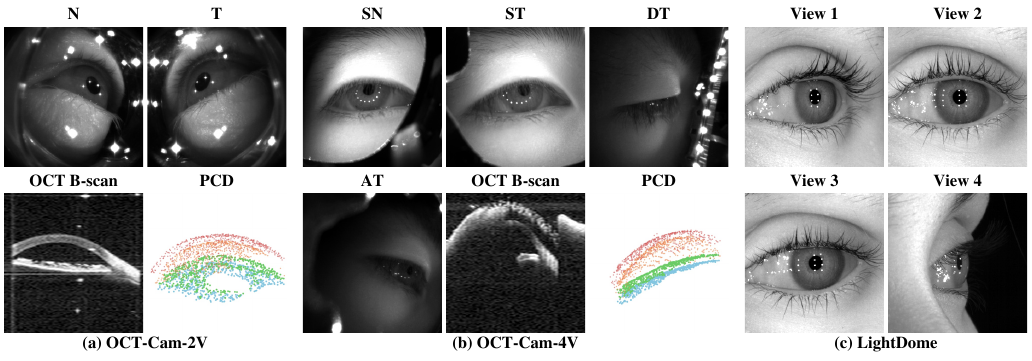}
    \caption{One example of \dsTwoV{}, \dsFourV{} and \dsMasq{}~\cite{Lin2026EyeTracker}: camera views, and a central OCT B-scan and the anatomy PCD where available.}
    \label{fig:s1_dataset_examples}
\end{figure}

\begin{figure}[H]
    \centering
    \includegraphics[width=\textwidth]{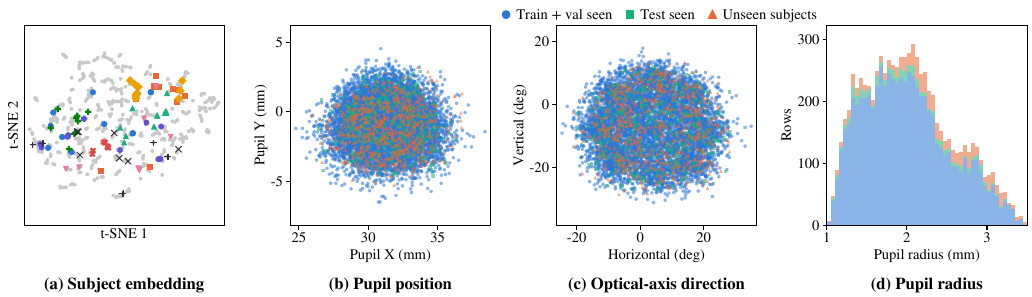}
    \caption{Subject embedding and label distributions of \dsTwoV{}.}
    \label{fig:embedding_octcam2v}
\end{figure}

\begin{figure}[H]
    \centering
    \includegraphics[width=\textwidth]{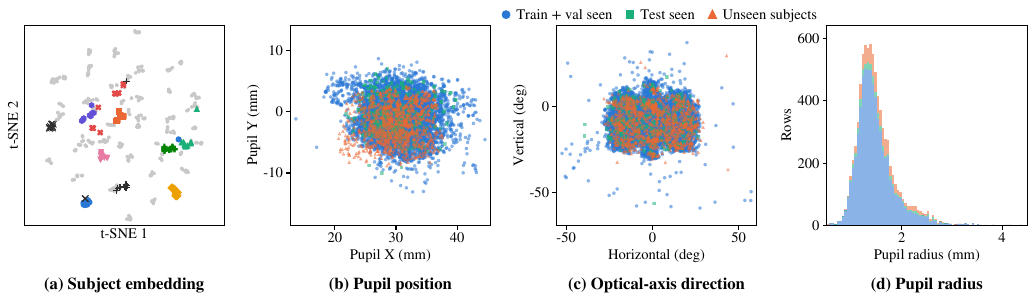}
    \caption{Subject embedding and label distributions of \dsFourV{}.}
    \label{fig:embedding_octcam4v}
\end{figure}

\noindent \textbf{Data Description} Tab.~\ref{tab:supp_datasets} lists the size, modalities and splits of each dataset, Fig.~\ref{fig:s1_dataset_examples} shows examples, and Figs.~\ref{fig:embedding_octcam2v} and~\ref{fig:embedding_octcam4v} show the subject embedding and label distributions. Test-seen holds samples of training subjects that are not used for training, and test-unseen holds subjects not seen in training. The light dome of \dsMasq{}~\cite{Lin2026EyeTracker} has 20 cameras, and we use its 18 close-up cameras and leave out the two wide face cameras. The labels of \dsMasq{}~\cite{Lin2026EyeTracker} and \dsNymeria{} are gaze directions (visual axis). For these two datasets, errors are reported in their gaze label space.

\begin{table}[H]\centering\caption{The datasets. Samples are OCT volumes (\dsTwoV{}, \dsFourV{}) or image frames (\dsMasq{}).}\label{tab:supp_datasets}
\fitbox{\textwidth}{\begin{tabular}{lrrrrlrrr}\toprule
Dataset & Subjects & Recordings & Samples & Views & Modalities & Train & Test-seen & Test-unseen \\ \midrule
\dsTwoV{} & 78 & 653 & 7931 & 2 & Images + OCT & 6331 & 829 & 771 \\
\dsFourV{} & 50 & 437 & 8265 & 4 & Images + OCT & 6460 & 800 & 1005 \\
\dsMasq{}~\cite{Lin2026EyeTracker} & 42 & -- & 204 & 18 & Images & 135 & -- & 69 \\
\bottomrule\end{tabular}}\end{table}

\FloatBarrier
\section{Implementation Details}
\label{sec:supp_impl}

\noindent \textbf{Architecture Details} All tokens have a dimension of $d=256$, and all attention layers use 8 heads and an MLP of width 1024. The OCT encoder takes the $D_o{\times}H_o{\times}W_o=100{\times}64{\times}128$ volume and applies three 3D convolution blocks with 64, 128 and 256 channels, the first two with a stride of 2. These features lie on the scan grid of the OCT, and we resample them onto a metric 3D grid using the OCT device calibration (OCT device warping). A strided 3D convolution then reduces them to the latent grid of $G=8{\times}15{\times}8=960$ cells, each covering about $3.1{\times}2.4{\times}2.1$\,mm. The PCD encoder embeds the 512 points of each label with a two-layer MLP and pools them into $K=8$ tokens per label by multi-head attention from learnable queries. The image encoder splits each view into $T=256$ patches of $16{\times}16$ pixels. The fusion transformer has $G$ learnable latent queries, one per cell, which cross-attend to all $G+4K+N_vT$ modality tokens in two layers and then self-attend in two layers. Linear heads then give the mean and log-variance of $z_{\mathrm{fused}}$. On the decoder side, the OCT decoder follows the reverse path. Two 3D convolution blocks upsample $z_{\mathrm{fused}}$ to 3D features $F$ on a $40{\times}60{\times}32$ metric grid, and after one more block, the features are resampled back onto the scan grid by the OCT device warping, where a last block gives $\hat{O}$. The PCD decoder maps 256 anatomy queries, 64 per label, through two cross-attention layers to 8 points each. Each camera-specific 2D decoder uses $16{\times}16$ queries per camera in two cross- and self-attention layers and upsamples the token grid to $256{\times}256$ with three 2D convolution blocks. The 3DGS decoder maps 64 Gaussian queries through four Transformer blocks to 1024 Gaussians each, which are rendered at $256{\times}256$ with 3D Gaussian splatting~\cite{Kerbl2023GS} using each rig's calibrated camera model. The DiT has 8 Transformer blocks of width 512, and the context latent enters it through a zero-initialized projection. The VAE and the DiT have 27.3\,M and 46.9\,M parameters, and Tab.~\ref{tab:supp_arch} summarizes the encoders and decoders.

\begin{table}[H]
    \centering
    \caption{Encoders and decoders of the VAE.}
    \label{tab:supp_arch}
    \footnotesize
    \begin{tabular}{lll}
    \toprule
    Encoder & Layers & Output \\
    \midrule
    OCT & 3 conv blocks, warping, strided conv & $G=960$ tokens \\
    PCD & MLP, attention pooling & $4K=32$ tokens \\
    Image & linear patch + Plücker ray embedding & $T=256$ tokens/view \\
    Fusion & 2 cross-attn, 2 self-attn & $G{\times}d=960{\times}256$ \\
    \bottomrule
    \end{tabular}

    \medskip
    \begin{tabular}{lll}
    \toprule
    Decoder & Layers & Output \\
    \midrule
    OCT & 3 conv blocks, warping, 1 conv\ block & $100{\times}64{\times}128$ vol \\
    PCD & 2 cross-attn, 256 queries & 2048 points \\
    2D & 2 cross-/self-attn, 3 conv blocks & $256{\times}256$ image \\
    3DGS & 4 Transformer blocks, 64 queries & 65{,}536 Gaussians \\
    \bottomrule
    \end{tabular}
\end{table}

\noindent \textbf{Training Details} The VAE is trained on the raw inputs of the training splits of \dsTwoV{}, \dsFourV{} and \dsMasq{} (only its image path for \dsMasq{}) for 60k iterations with a batch size of 4. In $\mathcal{L}_{\mathrm{VAE}}$ (Eq.~\ref{eq:vae_loss}), the LPIPS terms are weighted by 2 and $\mathcal{L}_{\mathrm{KL}}$ by $10^{-3}$. To support any number of input views, the VAE encoder receives a random subset of one to $N_v{-}1$ of the $N_v$ views with probability 0.5. In addition, the camera ray of each input view is replaced by a learned null ray with probability 0.25. The DiT is trained on the cached latents of \dsTwoV{} and \dsFourV{} for 20k iterations with a batch size of 128. It predicts the clean latent with velocity weighting, over 1{,}000 training timesteps on a cosine noise schedule shifted toward high noise, and samples are drawn with DDIM~\cite{Song2021DDIM} ($\eta{=}0$) in 50 steps. Training takes about 12 hours for the VAE and 1.6 hours for the DiT on one H100 GPU. Generating all outputs of one sample takes 1.1\,s on \dsTwoV{} and 1.8\,s on \dsFourV{} on one A100 at batch 1, with 1.2\,GB of memory.

\noindent \textbf{Condition Dropout} In the DiT, a context is given with probability 0.7, and the tone is always provided. The subject is dropped with probability 0.7 when a context is given and 0.3 otherwise, so a subject can be represented by its embedding, by its context images or by both. For classifier-free guidance~\cite{Ho2022CFG}, pupil position, pupil radius, optical axis and subject are dropped together with probability 0.1, and the guidance is applied at sampling on the predicted clean latent.

\noindent \textbf{Lighting Tone} The near-eye illumination and exposure differ between recordings, and we summarize them by a scalar tone $\tau$. For each sample, we take the median, over up to four other frames of the same recording, of the mean image intensity over all views, and standardize it with the mean and standard deviation of the training split. For test-seen samples, $\tau$ is read from training frames of the same recording. For test-unseen subjects, it is estimated from the given context images through a per-camera linear intensity response fitted on the training split, and it is set to the training mean ($\tau=0$) when no context is given.

\noindent \textbf{Biometrics Reading} We read the biometrics from an anatomy point cloud by RANSAC model fitting~\cite{Fischler1981RANSAC}. A sphere is fitted to each cornea surface from 4-point samples (99.999\% confidence, at most 10{,}000 iterations, inlier thresholds of 0.735 and 2.84\,mm for the front and back surfaces) and refitted on its inliers by robust least squares, which gives the corneal radii. A plane is fitted to each iris surface from 3-point samples (at most 10 iterations, inlier thresholds of 5.0 and 1.455\,mm for the front and back surfaces) and refitted on its inliers by PCA. For the pupil, we project the front and back iris points onto the back iris plane, take the innermost point in each angular bin as the pupil margin, and fit a circle to it, which gives the pupil center and radius. The optical axis points from the front-cornea sphere center to the pupil center. Along the normal of the front iris plane, the anterior chamber depth (ACD) is the distance from the back-cornea apex to the front iris plane, and the iris thickness is the distance between the two iris planes.

\noindent \textbf{Image Anatomy Readout} For outputs without an anatomy point cloud, we can read the pupil from the two near-eye views by stereo. A segmentation network gives the pupil region in each view, and an ellipse is fitted to its contour. The cornea is modeled as a sphere of radius 7.9\,mm with an effective refractive index of 1.33, and the pupil lies on a plane 3.5\,mm behind it, with a 2\,mm pupil radius as prior. The 3D pupil center and the cornea pose are fitted so that the pupil-contour rays of both views, refracted at the cornea, agree on the pupil plane. The pupil radius is the median distance from the fitted center to the refracted contour points on that plane, and the result is mapped to the OCT frame with a fixed camera-to-OCT offset. As the cornea shape and the pupil depth are fixed, this readout gives no subject-specific biometrics.

\FloatBarrier
\section{VAE Reconstruction Quality}
\label{sec:supp_recon}
Tab.~\ref{tab:s2_recon_table} reports the VAE reconstruction errors and Tab.~\ref{tab:s2_recon_anatomy} its anatomy biometric errors, and Figs.~\ref{fig:s2_recon_grid_a} and~\ref{fig:s2_recon_grid_b} show examples.

\begin{table}[H]\centering\caption{VAE reconstruction errors per dataset and split. Image metrics are over all views.}\label{tab:s2_recon_table}
\fitbox{\textwidth}{%
    \begin{tabular}{llcccccccccc}
    \toprule
    Dataset & Split & N & Pupil pos. (mm) $\downarrow$ & OA (\textdegree) & Pupil rad. (mm) $\downarrow$ & OCT PSNR (dB) $\uparrow$ & OCT LPIPS $\downarrow$ & Image PSNR (dB) $\uparrow$ & Image LPIPS $\downarrow$ & PCD Acc. (mm) $\downarrow$ & PCD Comp. (mm) $\downarrow$ \\
    \midrule
    \multirow{3}{*}{\dsTwoV{}} & Train & 512 & 0.109 & 1.01 & 0.074 & 25.25 & 0.190 & 26.52 & 0.065 & 0.075 & 0.171 \\
     & Test-seen & 380 & 0.106 & 0.87 & 0.074 & 25.30 & 0.189 & 26.30 & 0.077 & 0.072 & 0.171 \\
     & Test-unseen & 376 & 0.102 & 0.93 & 0.069 & 25.08 & 0.191 & 25.82 & 0.101 & 0.081 & 0.176 \\
    \midrule
    \multirow{3}{*}{\dsFourV{}} & Test-seen & 401 & 0.206 & 2.08 & 0.114 & 24.99 & 0.184 & 26.92 & 0.072 & 0.110 & 0.207 \\
     & Test-unseen & 601 & 0.139 & 1.16 & 0.098 & 24.40 & 0.178 & 24.72 & 0.135 & 0.102 & 0.198 \\
     & Train & 512 & 0.218 & 1.86 & 0.120 & 25.17 & 0.176 & 28.16 & 0.056 & 0.104 & 0.199 \\
    \bottomrule
    \end{tabular}%
}%
\end{table}
\begin{table}[H]\centering\caption{Anatomy biometric errors of the VAE reconstruction per dataset and split.}\label{tab:s2_recon_anatomy}
\fitbox{0.75\textwidth}{%
    \begin{tabular}{llccccc}
    \toprule
    Dataset & Split & N & Cornea front $R$ (mm) $\downarrow$ & Cornea back $R$ (mm) $\downarrow$ & ACD (mm) $\downarrow$ & Iris thick. (mm) $\downarrow$ \\
    \midrule
    \multirow{3}{*}{\dsTwoV{}} & Train & 512 & 0.130 & 0.124 & 0.040 & 0.031 \\
     & Test-seen & 380 & 0.128 & 0.116 & 0.038 & 0.027 \\
     & Test-unseen & 376 & 0.118 & 0.110 & 0.036 & 0.039 \\
    \midrule
    \multirow{3}{*}{\dsFourV{}} & Test-seen & 401 & 0.170 & 0.141 & 0.047 & 0.028 \\
     & Test-unseen & 601 & 0.182 & 0.168 & 0.052 & 0.030 \\
     & Train & 512 & 0.188 & 0.159 & 0.044 & 0.027 \\
    \bottomrule
    \end{tabular}%
}%
\end{table}

\suppfig{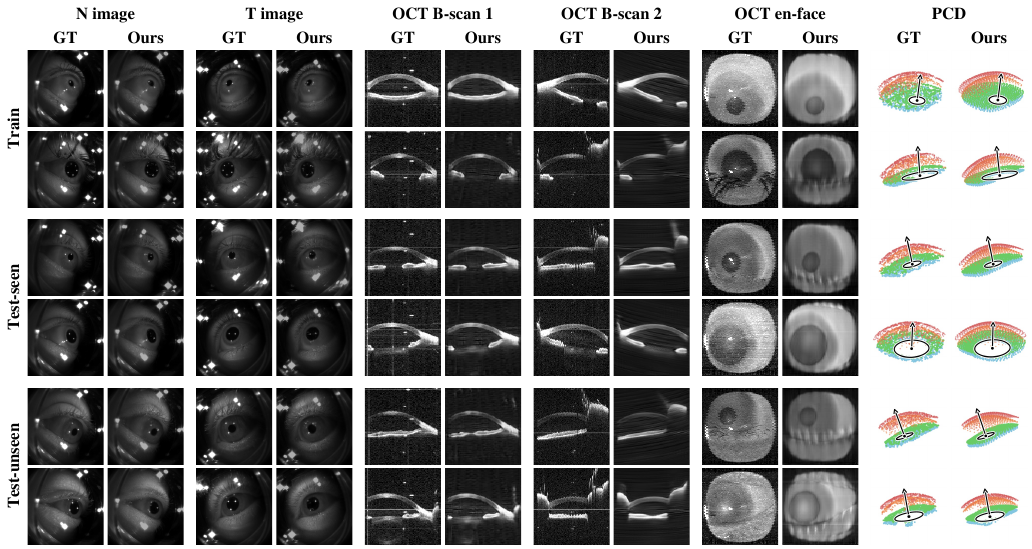}{Multi-modal reconstruction on \dsTwoV{}.}
\suppfig{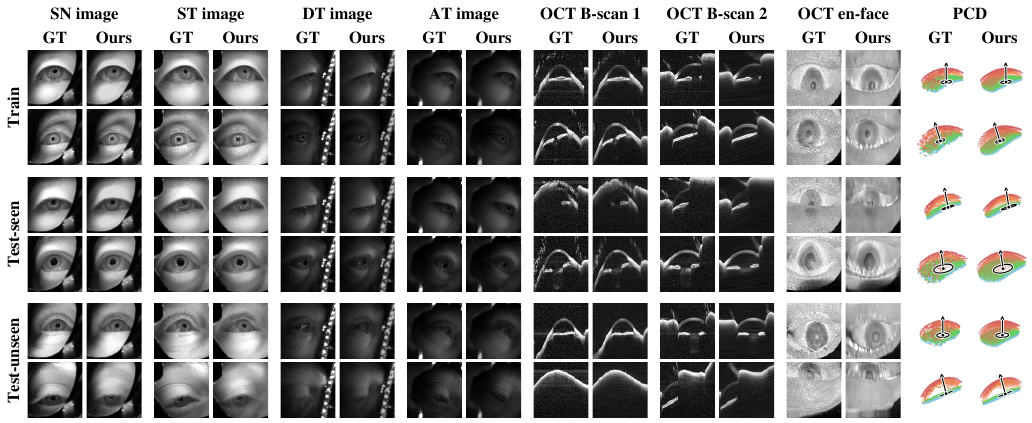}{Multi-modal reconstruction on \dsFourV{}.}

\FloatBarrier
\section{Conditional DiT Generation Quality}
\label{sec:supp_gen}
Tab.~\ref{tab:s3_gen_table} reports the generation errors and Tab.~\ref{tab:s3_gen_anatomy} its anatomy biometric errors, and Figs.~\ref{fig:s3_gen_grid_a} and~\ref{fig:s3_gen_grid_b} show generated samples.

\begin{table}[H]\centering\caption{Generation errors per dataset and split. Train and test-seen samples are generated from the subject embedding without context, and test-unseen samples from the nasal image of the same frame. Image metrics are over all views, including the context view.}\label{tab:s3_gen_table}
\fitbox{\textwidth}{%
    \begin{tabular}{llcccccccccc}
    \toprule
    Dataset & Split & N & Pupil pos. (mm) $\downarrow$ & OA (\textdegree) & Pupil rad. (mm) $\downarrow$ & OCT PSNR (dB) $\uparrow$ & OCT LPIPS $\downarrow$ & Image PSNR (dB) $\uparrow$ & Image LPIPS $\downarrow$ & PCD Acc. (mm) $\downarrow$ & PCD Comp. (mm) $\downarrow$ \\
    \midrule
    \multirow{3}{*}{\dsTwoV{}} & Train & 512 & 0.119 & 1.24 & 0.100 & 22.64 & 0.230 & 22.26 & 0.156 & 0.148 & 0.211 \\
     & Test-seen & 380 & 0.110 & 1.10 & 0.086 & 22.69 & 0.228 & 21.88 & 0.166 & 0.135 & 0.208 \\
     & Test-unseen & 376 & 0.139 & 1.11 & 0.139 & 21.64 & 0.247 & 25.40 & 0.106 & 0.223 & 0.259 \\
    \midrule
    \multirow{3}{*}{\dsFourV{}} & Test-seen & 401 & 0.159 & 0.83 & 0.143 & 21.46 & 0.232 & 25.49 & 0.084 & 0.192 & 0.276 \\
     & Test-unseen & 601 & 0.132 & 0.52 & 0.138 & 19.64 & 0.251 & 23.80 & 0.143 & 0.272 & 0.321 \\
     & Train & 512 & 0.169 & 0.92 & 0.141 & 21.55 & 0.220 & 26.65 & 0.069 & 0.181 & 0.270 \\
    \bottomrule
    \end{tabular}%
}%
\end{table}
\begin{table}[H]\centering\caption{Anatomy biometric errors of the generation per dataset and split. Test-unseen subjects are given one context image. These biometrics are not given as conditions: front and back corneal radii, anterior chamber depth (ACD, to the iris) and iris thickness.}\label{tab:s3_gen_anatomy}
\fitbox{0.75\textwidth}{%
    \begin{tabular}{llccccc}
    \toprule
    Dataset & Split & N & Cornea front $R$ (mm) $\downarrow$ & Cornea back $R$ (mm) $\downarrow$ & ACD (mm) $\downarrow$ & Iris thick. (mm) $\downarrow$ \\
    \midrule
    \multirow{3}{*}{\dsTwoV{}} & Train & 512 & 0.181 & 0.178 & 0.091 & 0.033 \\
     & Test-seen & 380 & 0.179 & 0.164 & 0.081 & 0.029 \\
     & Test-unseen & 376 & 0.148 & 0.184 & 0.232 & 0.040 \\
    \midrule
    \multirow{3}{*}{\dsFourV{}} & Test-seen & 401 & 0.159 & 0.135 & 0.079 & 0.030 \\
     & Test-unseen & 601 & 0.364 & 0.285 & 0.280 & 0.043 \\
     & Train & 512 & 0.231 & 0.201 & 0.077 & 0.033 \\
    \bottomrule
    \end{tabular}%
}%
\end{table}
\suppfig{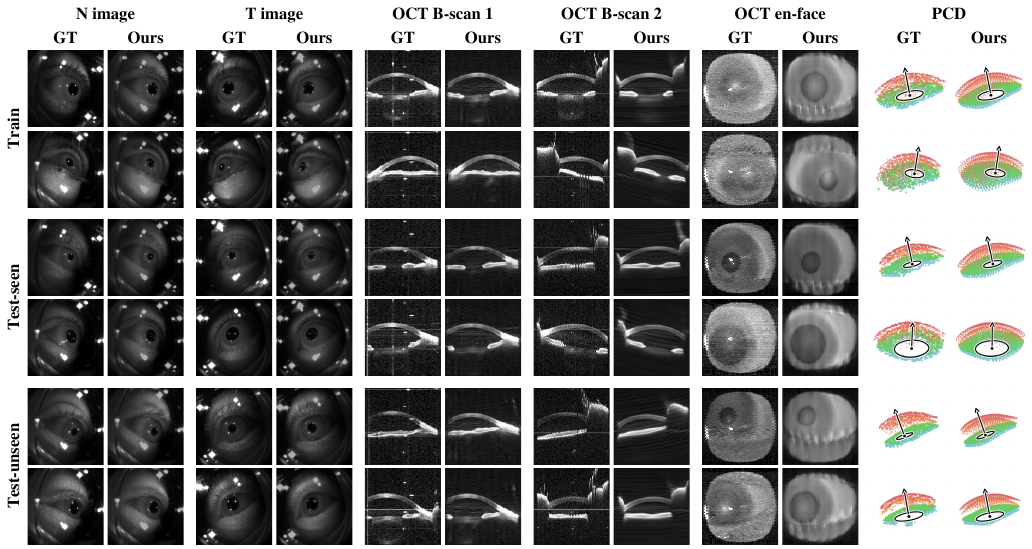}{Generated samples beside GT on \dsTwoV{}.}
\suppfig{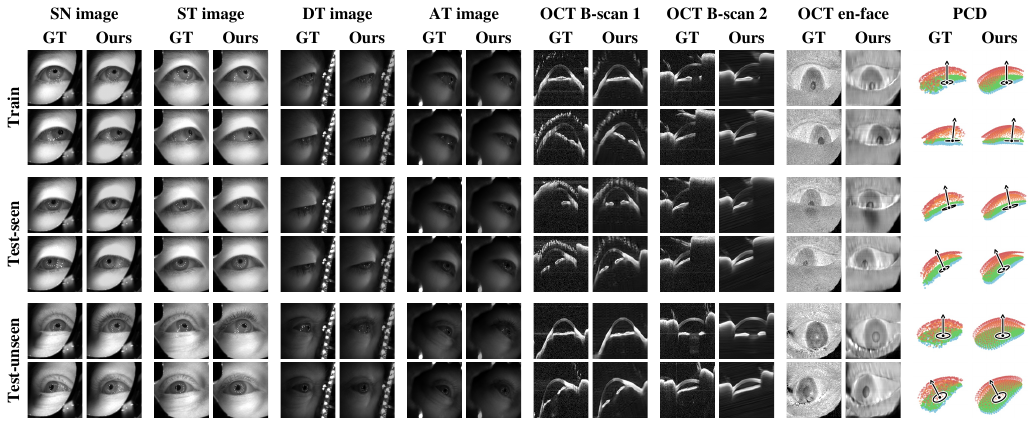}{Generated samples beside GT on \dsFourV{}.}

\FloatBarrier
\section{Cross-Modal Consistency}
\label{sec:supp_xmodal}
\begin{figure}[H]
    \centering
    \includegraphics[width=0.68\textwidth]{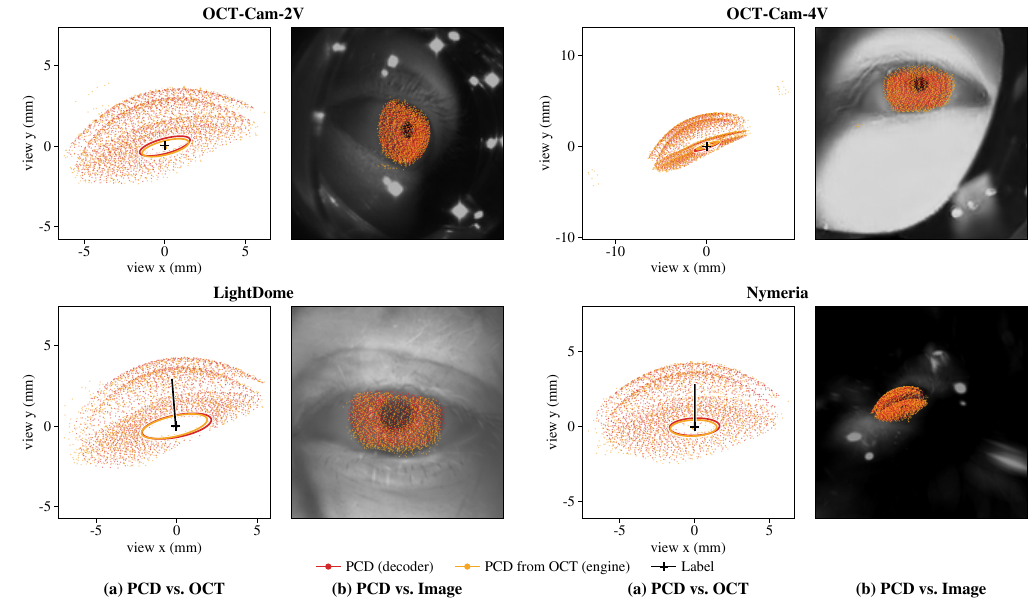}
    \caption{Cross-modal consistency, one sample per dataset.}
    \label{fig:s4_xmodal_grid}
\end{figure}

Tab.~\ref{tab:s4_xmodal_table} reports how well the pupil read from each generated modality agrees with the condition and with the other modalities on \dsTwoV{}, and Fig.~\ref{fig:s4_xmodal_grid} shows one sample per dataset.

\begin{table}[H]
    \centering
    \caption{Pupil consistency between the generated modalities on \dsTwoV{}. Command: error against the commanded pupil. Cross-Modal Consistency: error between two generated modalities.}
    \label{tab:s4_xmodal_table}
    {\small%
    \begin{tabular}{lcccccc}
    \toprule
     & \multicolumn{3}{c}{Command} & \multicolumn{3}{c}{Cross-Modal Consistency} \\
    \cmidrule(lr){2-4}
    \cmidrule(lr){5-7}
    Error $\downarrow$ & PCD & OCT & Image & OCT--PCD & Image--PCD & OCT--Image \\
    \midrule
    Pupil position (mm) & 0.086 & 0.159 & 0.473 & 0.146 & 0.467 & 0.510 \\
    Pupil radius (mm) & 0.091 & 0.112 & 0.160 & 0.091 & 0.124 & 0.149 \\
    \bottomrule
    \end{tabular}%
    }%
\end{table}

\FloatBarrier
\section{Novel-View Rendering}
\label{sec:supp_nv}
\begin{figure}[!ht]
    \centering
    \includegraphics[width=0.85\textwidth]{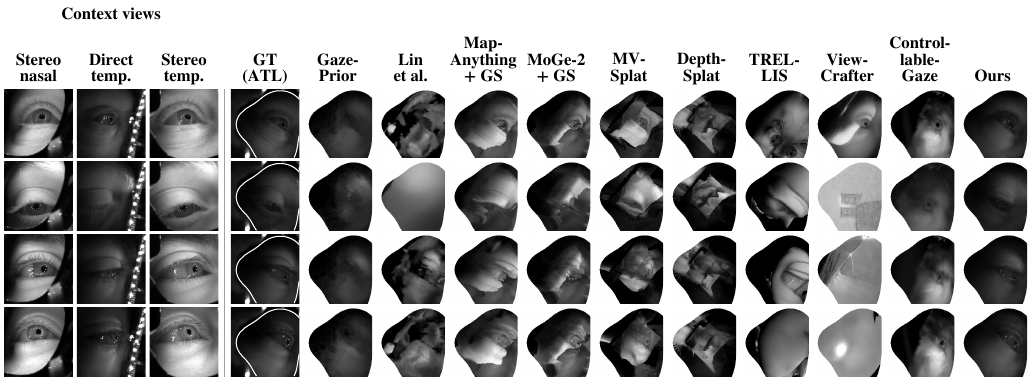}
    \caption{Novel-view rendering on \dsFourV{} test-unseen samples from three context views (left of the gray rule).}
    \label{fig:nv_grid_octcam4v}
\end{figure}

\begin{figure}[!ht]
    \centering
    \includegraphics[width=0.75\textwidth]{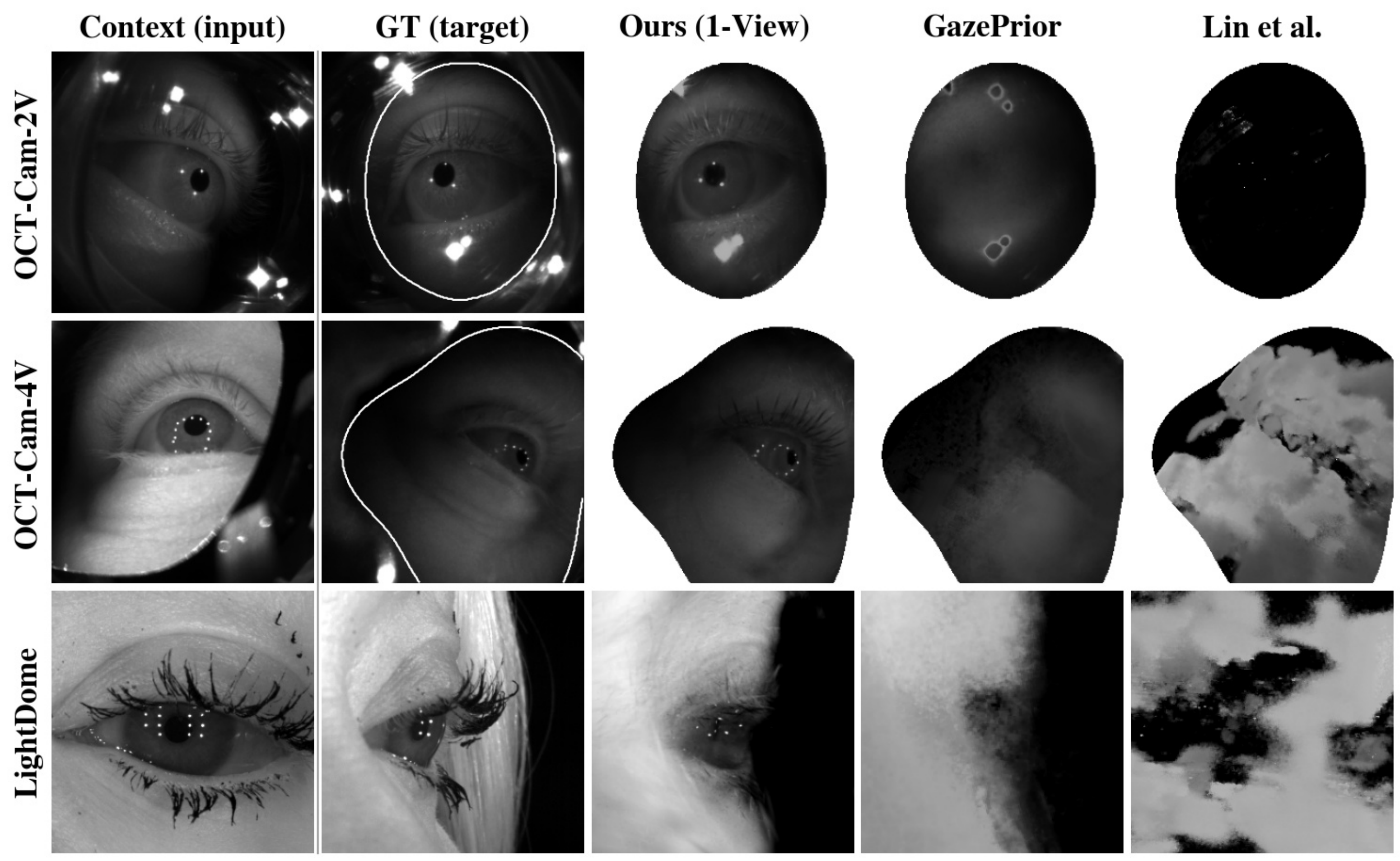}
    \caption{Novel-view rendering from one context image on \dsTwoV{}, \dsFourV{} and \dsMasq{}~\cite{Lin2026EyeTracker}.}
    \label{fig:qual_nv_4ds}
\end{figure}

We report novel-view rendering on \dsTwoV{} (Tab.~\ref{tab:nv_octcam2v}), \dsFourV{} (Tab.~\ref{tab:nv_octcam4v}) and \dsMasq{}~\cite{Lin2026EyeTracker} (Tab.~\ref{tab:s7_nv_table_lightdome}), followed by qualitative results (Figs.~\ref{fig:s7_nv_grid_a}, \ref{fig:nv_grid_octcam4v} and~\ref{fig:qual_nv_4ds}).

\noindent \textbf{Evaluation Views} On \dsTwoV{}, one view is given as context and the other is rendered and scored, in both directions. On \dsFourV{}, the AT view is rendered and scored from one (SN), two (SN, DT) or three (SN, DT, ST) context views. On \dsMasq{}, camera 9 is rendered and scored. Tab.~\ref{tab:s7_nv_perview_4v} scores every \dsFourV{} view from the SN context: the AT column matches Tab.~\ref{tab:nv_merged}, and the mean is a pooled average, with PSNR computed from the squared error of all four views of each sample.

\begin{figure}[H]
    \centering
    \includegraphics[width=0.6\textwidth]{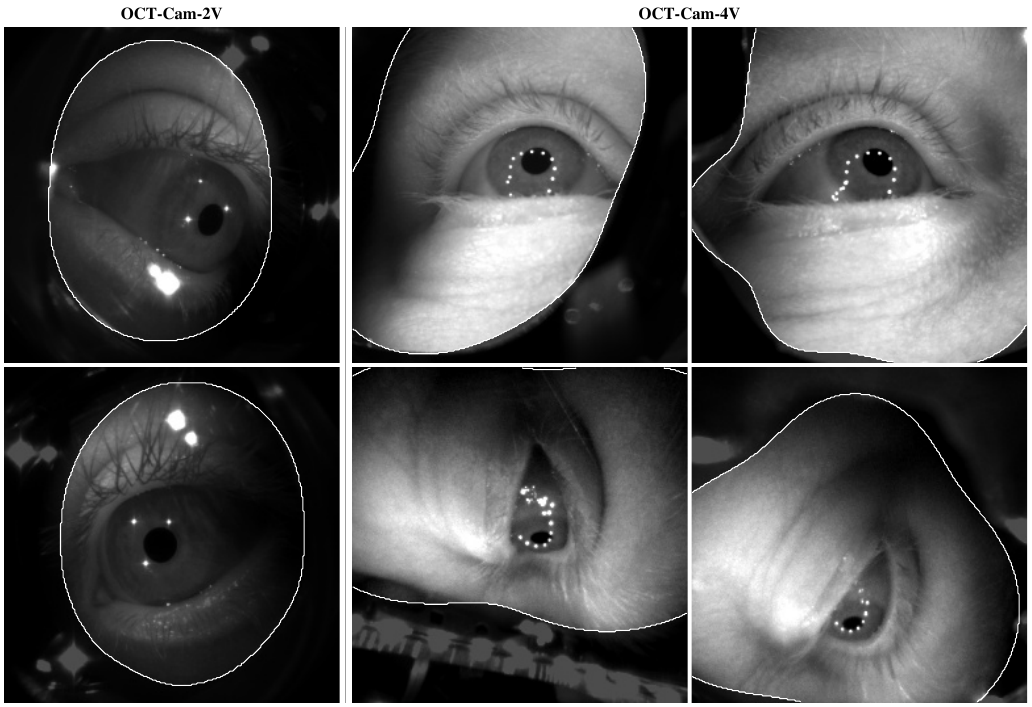}
    \caption{Field-of-view masks of the \dsTwoV{} (left) and \dsFourV{} (right) cameras.}
    \label{fig:s7_fov_masks}
\end{figure}

\noindent \textbf{FoV Mask} The measurement device blocks part of each camera's view, so the valid image area differs between cameras. We manually annotated a field-of-view mask for every camera of \dsTwoV{} and \dsFourV{} (Fig.~\ref{fig:s7_fov_masks}), and novel-view metrics on these datasets are computed inside this mask.

\begin{table}[H]
    \centering
    \caption{Per-view novel-view rendering on \dsFourV{} test-unseen subjects from one context view (SN), scored inside the FoV mask. AT is the target view of Tab.~\ref{tab:nv_merged}. The mean pools the masked error of all four views per sample (SSIM and LPIPS averaged over the views). Lin et al.\ fits each scene by optimization and is not re-rendered at the context view (--).}
    \label{tab:s7_nv_perview_4v}
    \fitbox{\textwidth}{%
    \begin{tabular}{lccccccccccccccc}
    \toprule
    & \multicolumn{3}{c}{SN (Context)} & \multicolumn{3}{c}{ST} & \multicolumn{3}{c}{DT} & \multicolumn{3}{c}{AT} & \multicolumn{3}{c}{Mean (all views)} \\
    \cmidrule(lr){2-4} \cmidrule(lr){5-7} \cmidrule(lr){8-10} \cmidrule(lr){11-13} \cmidrule(lr){14-16}
    Method & PSNR $\uparrow$ & SSIM $\uparrow$ & LPIPS $\downarrow$ & PSNR $\uparrow$ & SSIM $\uparrow$ & LPIPS $\downarrow$ & PSNR $\uparrow$ & SSIM $\uparrow$ & LPIPS $\downarrow$ & PSNR $\uparrow$ & SSIM $\uparrow$ & LPIPS $\downarrow$ & PSNR $\uparrow$ & SSIM $\uparrow$ & LPIPS $\downarrow$ \\
    \midrule
    Lin et al.~\cite{Lin2026EyeTracker} & -- & -- & -- & 13.38 & 0.463 & 0.588 & 11.72 & 0.404 & 0.621 & 11.77 & 0.375 & 0.665 & -- & -- & -- \\
    ControllableGaze~\cite{Li2025ControllableGaze} & 16.68 & 0.619 & 0.519 & 15.89 & 0.607 & 0.581 & 20.62 & 0.746 & 0.428 & 19.24 & 0.726 & 0.463 & 17.32 & 0.674 & 0.498 \\
    GazePrior~\cite{Dumery2026GazePrior} & 20.47 & 0.633 & 0.477 & 20.25 & 0.654 & 0.527 & 26.30 & 0.826 & 0.430 & 27.13 & 0.784 & 0.502 & 22.34 & 0.724 & 0.484 \\
    \midrule
    Ours & 20.44 & 0.641 & 0.216 & 20.63 & 0.645 & 0.234 & 26.58 & 0.853 & 0.173 & 29.23 & 0.891 & 0.137 & 22.67 & 0.757 & 0.190 \\
    \bottomrule
    \end{tabular}%
    }
\end{table}

\begin{table}[H]
    \centering
    \caption{One-view novel-view rendering on \dsTwoV{} test-seen samples.}
    \label{tab:nv_octcam2v}
    \begin{minipage}[t]{0.62\textwidth}
    \centering
    \fitbox{\linewidth}{%
        \begin{tabular}{lcccccc}
        \toprule
         & \multicolumn{3}{c}{Nasal $\rightarrow$ Temporal} & \multicolumn{3}{c}{Temporal $\rightarrow$ Nasal} \\
        \cmidrule(lr){2-4}
        \cmidrule(lr){5-7}
        (a) One-view & PSNR $\uparrow$ & SSIM $\uparrow$ & LPIPS $\downarrow$ & PSNR $\uparrow$ & SSIM $\uparrow$ & LPIPS $\downarrow$ \\
        \midrule
        GazePrior~\cite{Dumery2026GazePrior} & 20.23 & 0.694 & 0.468 & 20.31 & \textit{0.676} & 0.485 \\
        Lin et al.~\cite{Lin2026EyeTracker} & 14.15 & 0.267 & 0.595 & 12.49 & 0.307 & 0.603 \\
        ControllableGaze~\cite{Li2025ControllableGaze} & \textit{21.81} & \textit{0.702} & 0.397 & \textit{20.98} & 0.671 & 0.408 \\
        \textbf{Ours} & \textbf{24.89} & \textbf{0.786} & \textbf{0.128} & \textbf{24.60} & \textbf{0.762} & \textbf{0.146} \\
        \midrule
        MapAnything~\cite{Keetha2026MapAnything} + GS Head & 19.62 & 0.653 & \textit{0.355} & 19.74 & 0.635 & \textit{0.364} \\
        MoGe-2~\cite{Wang2025MoGe2} + GS Head & 19.10 & 0.630 & 0.395 & 19.18 & 0.613 & 0.377 \\
        DA-V2~\cite{Yang2024DepthAnythingV2} + GS Head & 18.82 & 0.634 & 0.439 & 17.44 & 0.582 & 0.539 \\
        \midrule
        TRELLIS~\cite{Xiang2025TRELLIS} & 15.20 & 0.514 & 0.581 & 16.32 & 0.518 & 0.538 \\
        ViewCrafter~\cite{Yu2025ViewCrafter} & 15.56 & 0.501 & 0.514 & 12.75 & 0.404 & 0.539 \\
        \bottomrule
        \end{tabular}%
    }
    \end{minipage}%
\end{table}
\begin{table}[H]
    \centering
    \caption{Novel-view rendering on \dsFourV{} test-unseen samples from one to three context views. $\times$: the method needs at least two views.}
    \label{tab:nv_octcam4v}
    {\footnotesize%
    \fitbox{\textwidth}{%
        \begin{tabular}{lccccccccc}
        \toprule
         & \multicolumn{3}{c}{1 context view} & \multicolumn{3}{c}{2 context views} & \multicolumn{3}{c}{3 context views} \\
        \cmidrule(lr){2-4}
        \cmidrule(lr){5-7}
        \cmidrule(lr){8-10}
        Method & PSNR $\uparrow$ & SSIM $\uparrow$ & LPIPS $\downarrow$ & PSNR $\uparrow$ & SSIM $\uparrow$ & LPIPS $\downarrow$ & PSNR $\uparrow$ & SSIM $\uparrow$ & LPIPS $\downarrow$ \\
        \midrule
        \textbf{Ours} & \textbf{29.23} & \textbf{0.891} & \textbf{0.137} & \textbf{29.40} & \textbf{0.893} & \textbf{0.132} & \textbf{29.42} & \textbf{0.894} & \textbf{0.130} \\
        \midrule
        GazePrior~\cite{Dumery2026GazePrior} & \textit{27.13} & \textit{0.784} & 0.502 & \textit{27.20} & \textit{0.783} & 0.503 & \textit{27.54} & \textit{0.788} & 0.499 \\
        Lin et al.~\cite{Lin2026EyeTracker} & 11.77 & 0.375 & 0.665 & 17.59 & 0.519 & 0.600 & 15.85 & 0.480 & 0.582 \\
        ControllableGaze~\cite{Li2025ControllableGaze} & 19.24 & 0.726 & \textit{0.463} & 18.95 & 0.721 & 0.473 & 18.93 & 0.725 & \textit{0.466} \\
        \midrule
        MapAnything~\cite{Keetha2026MapAnything} + GS Head & 13.66 & 0.575 & 0.500 & 14.96 & 0.697 & \textit{0.463} & 14.50 & 0.681 & 0.481 \\
        MoGe-2~\cite{Wang2025MoGe2} + GS Head & 15.35 & 0.576 & 0.516 & 18.39 & 0.691 & 0.465 & 17.95 & 0.685 & 0.468 \\
        DA-V2~\cite{Yang2024DepthAnythingV2} + GS Head & 14.08 & 0.350 & 0.677 & 16.07 & 0.505 & 0.632 & 15.10 & 0.488 & 0.665 \\
        YoNoSplat~\cite{Ye2026YoNoSplat} & $\times$ & $\times$ & $\times$ & 18.73 & 0.578 & 0.472 & 14.67 & 0.689 & 0.535 \\
        DepthSplat~\cite{Xu2025DepthSplat} & $\times$ & $\times$ & $\times$ & 13.72 & 0.431 & 0.618 & 14.56 & 0.406 & 0.664 \\
        MVSplat~\cite{Chen2024MVSplat} & $\times$ & $\times$ & $\times$ & 17.73 & 0.554 & 0.566 & 14.34 & 0.490 & 0.621 \\
        ReSplat~\cite{Xu2026ReSplat} & $\times$ & $\times$ & $\times$ & 19.59 & 0.602 & 0.505 & 14.06 & 0.595 & 0.481 \\
        TRELLIS~\cite{Xiang2025TRELLIS} & 13.38 & 0.229 & 0.653 & 15.12 & 0.430 & 0.612 & 15.74 & 0.525 & 0.555 \\
        ViewCrafter~\cite{Yu2025ViewCrafter} & 8.07 & 0.439 & 0.711 & 15.39 & 0.630 & 0.515 & 13.14 & 0.514 & 0.676 \\
        \bottomrule
        \end{tabular}%
    }
    }%
\end{table}
\begin{table}[H]\centering\caption{Novel-view rendering on \dsMasq{}~\cite{Lin2026EyeTracker} test-unseen subjects, target camera 9.}\label{tab:s7_nv_table_lightdome}
{\footnotesize%
\begin{tabular}{lcccccc}
\toprule
 & \multicolumn{3}{c}{1 context view} & \multicolumn{3}{c}{3 context views} \\
\cmidrule(lr){2-4}
\cmidrule(lr){5-7}
Method & PSNR $\uparrow$ & SSIM $\uparrow$ & LPIPS $\downarrow$ & PSNR $\uparrow$ & SSIM $\uparrow$ & LPIPS $\downarrow$ \\
\midrule
GazePrior~\cite{Dumery2026GazePrior} & \textit{16.08} & \textit{0.485} & \textit{0.576} & \textit{17.27} & \textit{0.493} & \textit{0.554} \\
Lin et al.~\cite{Lin2026EyeTracker} & 11.73 & 0.404 & 0.723 & 14.22 & 0.405 & 0.588 \\
\midrule
\textbf{Ours} & \textbf{18.08} & \textbf{0.502} & \textbf{0.358} & \textbf{19.30} & \textbf{0.520} & \textbf{0.324} \\
\bottomrule
\end{tabular}%
}%
\end{table}
\suppfig{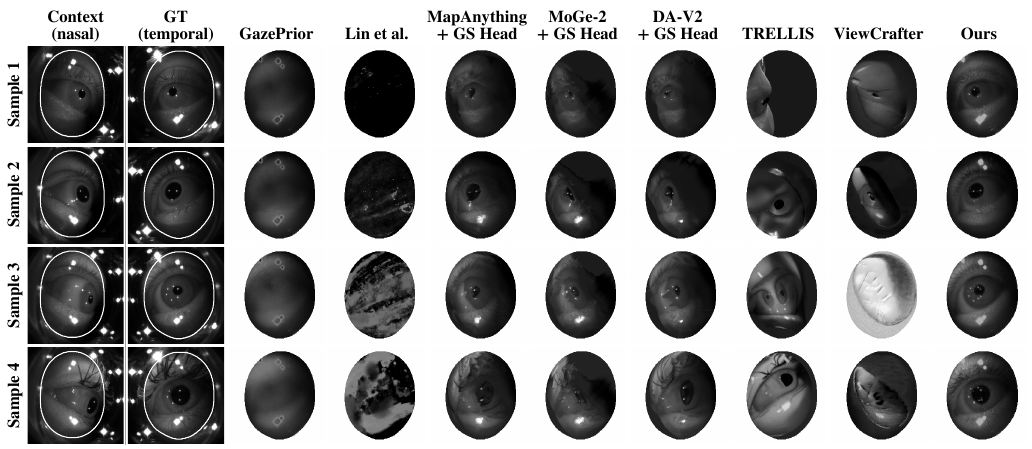}{Novel-view rendering on \dsTwoV{} test-seen samples.}

\FloatBarrier
\section{Number of Context Views}
\label{sec:supp_view_sparsification}
\begin{figure}[H]
    \centering
    \includegraphics[width=0.75\textwidth]{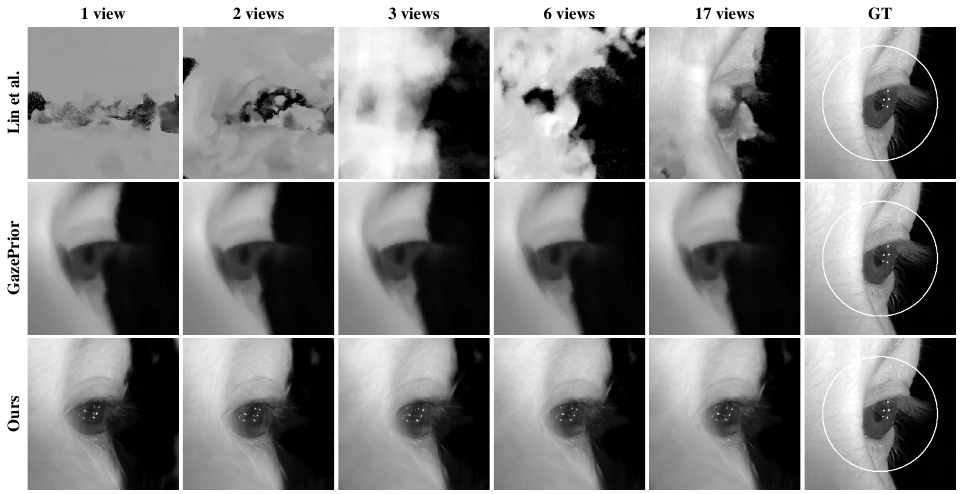}
    \caption{View sparsification on \dsMasq{}~\cite{Lin2026EyeTracker}, target camera 9.}
    \label{fig:qual_view_sparsification}
\end{figure}

Tab.~\ref{tab:view_sparsification} and Fig.~\ref{fig:qual_view_sparsification} report novel-view rendering on \dsMasq{}~\cite{Lin2026EyeTracker} with 1 to 17 context views and a held-out target camera.

\begin{table}[H]
    \centering
    \caption{View sparsification on \dsMasq{}~\cite{Lin2026EyeTracker}, target camera 9. No fit: one forward pass from the offered views. Test-time fit: optimization on the offered views.}
    \label{tab:view_sparsification}
    \fitbox{\textwidth}{%
        \begin{tabular}{lccccccccccccccc}
        \toprule
         & \multicolumn{3}{c}{1 view} & \multicolumn{3}{c}{2 views} & \multicolumn{3}{c}{3 views} & \multicolumn{3}{c}{6 views} & \multicolumn{3}{c}{17 views} \\
        \cmidrule(lr){2-4}
        \cmidrule(lr){5-7}
        \cmidrule(lr){8-10}
        \cmidrule(lr){11-13}
        \cmidrule(lr){14-16}
        Method & PSNR $\uparrow$ & SSIM $\uparrow$ & LPIPS $\downarrow$ & PSNR $\uparrow$ & SSIM $\uparrow$ & LPIPS $\downarrow$ & PSNR $\uparrow$ & SSIM $\uparrow$ & LPIPS $\downarrow$ & PSNR $\uparrow$ & SSIM $\uparrow$ & LPIPS $\downarrow$ & PSNR $\uparrow$ & SSIM $\uparrow$ & LPIPS $\downarrow$ \\
        \midrule
        Lin et al.~\cite{Lin2026EyeTracker} (per-sample fit) & 11.73 & 0.404 & 0.723 & 12.33 & 0.410 & 0.662 & 14.22 & 0.405 & 0.588 & 13.47 & 0.407 & 0.578 & 17.16 & 0.510 & 0.473 \\
        GazePrior~\cite{Dumery2026GazePrior} (no fit) & 15.97 & \textbf{0.548} & 0.564 & 15.97 & \textbf{0.548} & 0.564 & 15.97 & \textit{0.548} & 0.564 & 15.97 & \textit{0.548} & 0.564 & 15.97 & \textit{0.548} & 0.564 \\
        GazePrior~\cite{Dumery2026GazePrior} (test-time fit) & 16.08 & 0.485 & 0.576 & 16.90 & 0.494 & 0.563 & 17.27 & 0.493 & 0.554 & 17.37 & 0.499 & 0.559 & 17.34 & 0.499 & 0.565 \\
        \textbf{Ours} (no fit) & \textbf{18.08} & \textit{0.502} & \textbf{0.358} & \textbf{18.93} & \textit{0.516} & \textbf{0.329} & \textit{19.30} & 0.520 & \textit{0.324} & \textit{19.25} & 0.514 & \textit{0.329} & \textit{19.38} & 0.515 & \textit{0.321} \\
        \textbf{Ours} (test-time fit) & \textit{17.72} & 0.499 & \textit{0.387} & \textit{18.79} & \textit{0.516} & \textit{0.375} & \textbf{21.52} & \textbf{0.597} & \textbf{0.317} & \textbf{22.83} & \textbf{0.613} & \textbf{0.310} & \textbf{23.36} & \textbf{0.633} & \textbf{0.300} \\
        \bottomrule
        \end{tabular}%
    }%
\end{table}

\FloatBarrier
\section{Sim-to-Real Eye Tracking}
\label{sec:supp_et}

\begin{figure}[!ht]
    \centering
    \includegraphics[width=\textwidth]{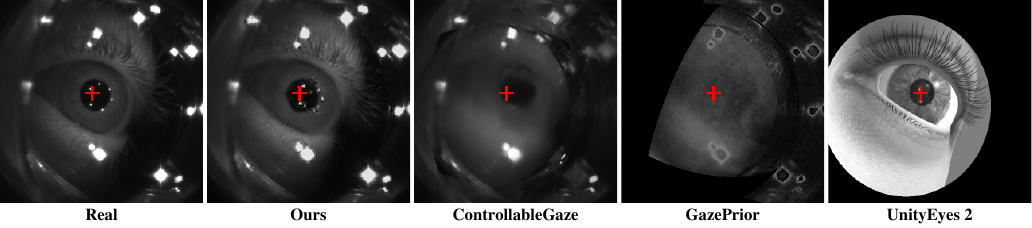}
    \caption{Training images of each sim-to-real source for the same condition (nasal camera). Red cross: the commanded pupil center projected with the rig calibration. GazePrior renders blurry images in which the pupil is barely visible, and UnityEyes 2 differs strongly from the IR images in appearance.}
    \label{fig:s2r_baseline_samples}
\end{figure}

\noindent \textbf{Protocol} A generator synthesizes the ET training images (Fig.~\ref{fig:s2r_baseline_samples}) from the conditions of real training frames. An ET estimator is trained on these synthetic images, or on the real training images as an oracle, and is evaluated on real test frames. On \dsTwoV{}, the volumes of each recording are split into source, train and test parts, and the generator is fitted on the source part, so no train or test frame is seen by the generator. On \dsNymeria{}, the test participants are held out from both the generator and the ET estimator. All training sources use the same model and training recipe, with three seeds, and Real + Ours is trained on the union of the real and our synthetic training images for the same number of epochs. Angular errors are reported after a per-subject bias calibration, as in the user calibration of ET systems: for each test subject, five evenly spaced test frames are used to estimate a constant yaw and pitch offset of the estimates, which is removed from the estimates on the other test frames, and these five frames are not scored.

\begin{table}[H]
    \centering
    \caption{Tab.~\ref{tab:sim_to_real_aria} with the half-width of the 95\% bootstrap confidence interval (seeds and test frames) after each value.}
    \label{tab:supp_s2r_ci}
    \fitbox{\textwidth}{%
        \begin{tabular}{lcccccccccccc}
        \toprule
         & \multicolumn{9}{c}{\dsTwoV{}} & \multicolumn{3}{c}{\dsNymeria{}} \\
        \cmidrule(lr){2-10} \cmidrule(lr){11-13}
         & \multicolumn{3}{c}{Pupil pos. (mm) $\downarrow$} & \multicolumn{3}{c}{Pupil rad. (mm) $\downarrow$} & \multicolumn{3}{c}{Optical axis ($^\circ$) $\downarrow$} & \multicolumn{3}{c}{Gaze ($^\circ$) $\downarrow$} \\
        \cmidrule(lr){2-4} \cmidrule(lr){5-7} \cmidrule(lr){8-10} \cmidrule(lr){11-13}
        ET trained on & E50 & E75 & E95 & E50 & E75 & E95 & E50 & E75 & E95 & E50 & E75 & E95 \\
        \midrule
        ControllableGaze~\cite{Li2025ControllableGaze} & \textit{3.20} {\scriptsize$\pm$0.25} & \textit{4.32} {\scriptsize$\pm$0.29} & \textit{6.44} {\scriptsize$\pm$0.46} & 0.65 {\scriptsize$\pm$0.19} & 1.02 {\scriptsize$\pm$0.23} & 1.49 {\scriptsize$\pm$0.25} & \textit{9.36} {\scriptsize$\pm$0.85} & \textit{12.70} {\scriptsize$\pm$0.92} & \textit{18.70} {\scriptsize$\pm$1.72} & $\times$ & $\times$ & $\times$ \\
        GazePrior~\cite{Dumery2026GazePrior} & 18.96 {\scriptsize$\pm$0.98} & 21.17 {\scriptsize$\pm$1.10} & 24.46 {\scriptsize$\pm$1.03} & 0.58 {\scriptsize$\pm$0.13} & 0.95 {\scriptsize$\pm$0.20} & 1.53 {\scriptsize$\pm$0.30} & 16.68 {\scriptsize$\pm$2.43} & 23.69 {\scriptsize$\pm$2.69} & 36.64 {\scriptsize$\pm$4.07} & \textit{8.41} {\scriptsize$\pm$0.88} & \textit{12.05} {\scriptsize$\pm$1.34} & \textit{18.89} {\scriptsize$\pm$2.61} \\
        UnityEyes 2~\cite{Smith2025UnityEyes2} & 10.55 {\scriptsize$\pm$0.83} & 12.37 {\scriptsize$\pm$0.88} & 16.69 {\scriptsize$\pm$1.18} & \textit{0.57} {\scriptsize$\pm$0.21} & \textit{0.90} {\scriptsize$\pm$0.23} & \textit{1.42} {\scriptsize$\pm$0.21} & 12.90 {\scriptsize$\pm$1.28} & 18.83 {\scriptsize$\pm$1.69} & 27.40 {\scriptsize$\pm$4.38} & 10.18 {\scriptsize$\pm$1.91} & 13.89 {\scriptsize$\pm$2.04} & 24.00 {\scriptsize$\pm$6.34} \\
        \textbf{Ours} & \textbf{0.61} {\scriptsize$\pm$0.09} & \textbf{0.85} {\scriptsize$\pm$0.12} & \textbf{1.33} {\scriptsize$\pm$0.17} & \textbf{0.34} {\scriptsize$\pm$0.33} & \textbf{0.43} {\scriptsize$\pm$0.33} & \textbf{0.59} {\scriptsize$\pm$0.31} & \textbf{4.17} {\scriptsize$\pm$0.35} & \textbf{6.01} {\scriptsize$\pm$0.48} & \textbf{9.16} {\scriptsize$\pm$0.85} & \textbf{3.53} {\scriptsize$\pm$0.47} & \textbf{6.18} {\scriptsize$\pm$1.06} & \textbf{14.91} {\scriptsize$\pm$5.39} \\
        \midrule
        Real (oracle) & 0.64 {\scriptsize$\pm$0.15} & 0.80 {\scriptsize$\pm$0.17} & 1.12 {\scriptsize$\pm$0.15} & 0.25 {\scriptsize$\pm$0.23} & 0.31 {\scriptsize$\pm$0.24} & 0.41 {\scriptsize$\pm$0.26} & 2.84 {\scriptsize$\pm$0.28} & 4.21 {\scriptsize$\pm$0.35} & 6.20 {\scriptsize$\pm$0.65} & 2.59 {\scriptsize$\pm$0.45} & 4.17 {\scriptsize$\pm$0.75} & 8.50 {\scriptsize$\pm$2.52} \\
        Real + Ours & 0.44 {\scriptsize$\pm$0.02} & 0.58 {\scriptsize$\pm$0.02} & 0.90 {\scriptsize$\pm$0.08} & 0.10 {\scriptsize$\pm$0.05} & 0.15 {\scriptsize$\pm$0.06} & 0.25 {\scriptsize$\pm$0.05} & 2.75 {\scriptsize$\pm$0.27} & 4.29 {\scriptsize$\pm$0.40} & 6.91 {\scriptsize$\pm$0.96} & 2.28 {\scriptsize$\pm$0.36} & 3.48 {\scriptsize$\pm$0.55} & 7.47 {\scriptsize$\pm$1.96} \\
        \bottomrule
        \end{tabular}%
    }%
\end{table}

\noindent \textbf{Seed Variance and Confidence Intervals} Tab.~\ref{tab:supp_s2r_ci} repeats Tab.~\ref{tab:sim_to_real_aria} with the half-width of the 95\% confidence interval of each value, obtained by resampling the three seeds and the test frames with replacement (2{,}000 resamples). The optical-axis E50 of our data is $4.17^\circ$ (95\% CI $[3.84, 4.54]^\circ$) on \dsTwoV{}, and the gaze E50 is $3.53^\circ$ (95\% CI $[3.17, 4.11]^\circ$) on \dsNymeria{}.

\begin{figure}[!ht]
    \centering
    \includegraphics[width=0.82\textwidth]{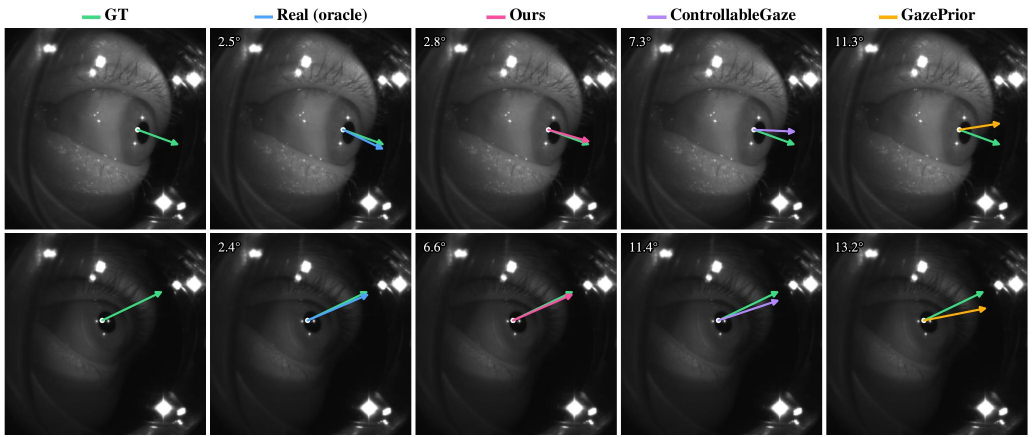}
    \caption{Eye-tracking estimates on real \dsTwoV{} test frames of two subjects, by training source. Green: label. Each panel shows the estimate and its optical-axis error.}
    \label{fig:et_qual_twov}
    \vspace{8pt}
    \includegraphics[width=0.92\textwidth]{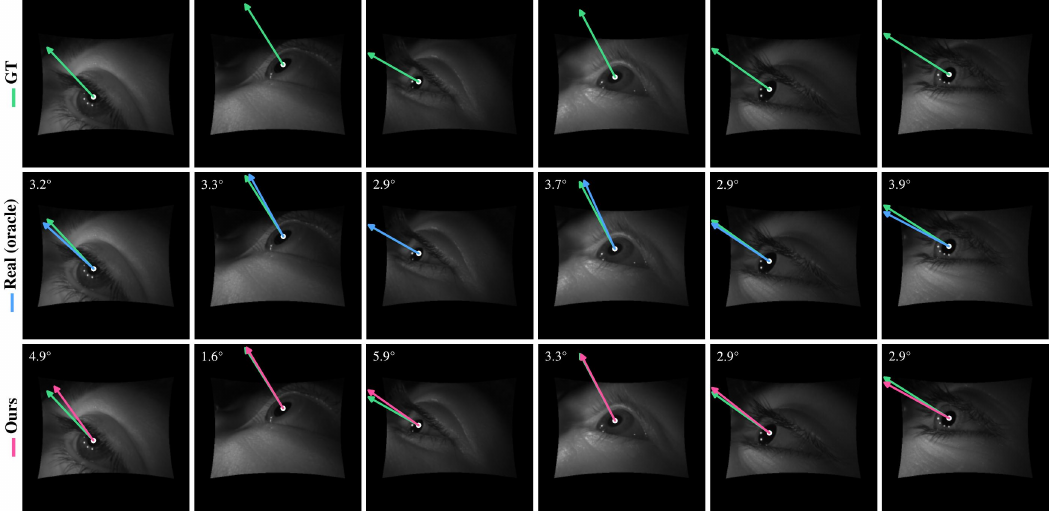}
    \caption{Gaze estimates on real \dsNymeria{} test frames of six participants, for estimators trained on real data and on Ours. Green: label. Each panel shows the estimate and its gaze error.}
    \label{fig:et_qual_nymeria}
\end{figure}

\noindent \textbf{Aria ET Model} We use the Aria ET model~\cite{Mansour2025AriaET}, which encodes the eye images with a lightweight ResNet backbone~\cite{He2016ResNet} and regresses the gaze direction as yaw and pitch angles. We train it from scratch for 50 epochs with Adam, a batch size of 64 and a learning rate of $5\times10^{-4}$, and use the final checkpoint. The original model predicts only the gaze. To also train it for the pupil position and radius on \dsTwoV{}, we encode each of the two views with a shared standard ResNet-18~\cite{He2016ResNet}, and attach a 3D gaze head, a 3D pupil-position head and a linear pupil-radius head, the last on the pooled features of both views. On \dsNymeria{}, which has only gaze labels, we train the original gaze output. Figs.~\ref{fig:et_qual_twov} and~\ref{fig:et_qual_nymeria} show the estimates on real test frames.
\FloatBarrier

\section{Novel-View Baselines Trained on OCT-Cam}
\label{sec:supp_nv_trained}

To test whether the general novel-view baselines are limited by the domain gap, we train MVSplat~\cite{Chen2024MVSplat} and DepthSplat~\cite{Xu2025DepthSplat} on the training split of \dsFourV{}, whose four cameras provide real images of held-out views for supervision. Each model starts from its released checkpoint, and all of its parameters are trained for 4,000 steps with a batch size of 4 and AdamW at a learning rate of $2\times10^{-5}$. At each step, two or three of the four views of a training frame are given as context, and the remaining view is rendered and compared with its real image by an L1 and LPIPS loss inside the FoV mask. We keep the checkpoint with the lowest validation LPIPS with three context views, and evaluate the trained models with the protocol of Tab.~\ref{tab:nv_octcam4v} on the test-unseen subjects.

As shown in Tab.~\ref{tab:supp_nv_trained}, training on \dsFourV{} raises the PSNR of both models by 6.5 to 18.2\,dB over their released checkpoints. With three context views, the trained models reach 32.49 and 32.00\,dB and exceed our model in PSNR and SSIM, while our model keeps the lowest LPIPS (0.130 against 0.204 and 0.182). With two context views, our model is better in all metrics (29.40 against 26.99\,dB for DepthSplat), and with one context view these feed-forward models cannot be applied. The large gain from the third view, which forms a stereo pair with the nasal view, suggests that the trained models mostly rely on multi-view matching of the given views, whereas our model renders the target from one view almost as well as from three (Fig.~\ref{fig:s7_nv_trained_baselines_ctx2}).

\begin{table}[!ht]
    \centering
    \caption{Novel-view rendering on test-unseen \dsFourV{} subjects of MVSplat and DepthSplat with their released checkpoints (Pretrained) and after training on \dsFourV{}, scored inside the FoV mask as in Tab.~\ref{tab:nv_octcam4v}. $\times$: the method needs at least two context views.}
    \label{tab:supp_nv_trained}
    {\footnotesize%
    \fitbox{\textwidth}{%
    \begin{tabular}{llccccccccc}
    \toprule
     & & \multicolumn{3}{c}{1 context view} & \multicolumn{3}{c}{2 context views} & \multicolumn{3}{c}{3 context views} \\
    \cmidrule(lr){3-5}
    \cmidrule(lr){6-8}
    \cmidrule(lr){9-11}
    Method & Training & PSNR $\uparrow$ & SSIM $\uparrow$ & LPIPS $\downarrow$ & PSNR $\uparrow$ & SSIM $\uparrow$ & LPIPS $\downarrow$ & PSNR $\uparrow$ & SSIM $\uparrow$ & LPIPS $\downarrow$ \\
    \midrule
    MVSplat & Pretrained & $\times$ & $\times$ & $\times$ & 17.73 & 0.554 & 0.566 & 14.34 & 0.490 & 0.621 \\
    MVSplat & \dsFourV{} & $\times$ & $\times$ & $\times$ & 24.27 & 0.801 & 0.299 & \textbf{32.49} & \textit{0.904} & 0.204 \\
    DepthSplat & Pretrained & $\times$ & $\times$ & $\times$ & 13.72 & 0.431 & 0.618 & 14.56 & 0.406 & 0.664 \\
    DepthSplat & \dsFourV{} & $\times$ & $\times$ & $\times$ & \textit{26.99} & \textit{0.875} & \textit{0.237} & \textit{32.00} & \textbf{0.907} & \textit{0.182} \\
    \midrule
    \textbf{Ours} & OCT-Cam & \textbf{29.23} & \textbf{0.891} & \textbf{0.137} & \textbf{29.40} & \textbf{0.893} & \textbf{0.132} & 29.42 & 0.894 & \textbf{0.130} \\
    \bottomrule
    \end{tabular}%
    }}
\end{table}

\suppfig{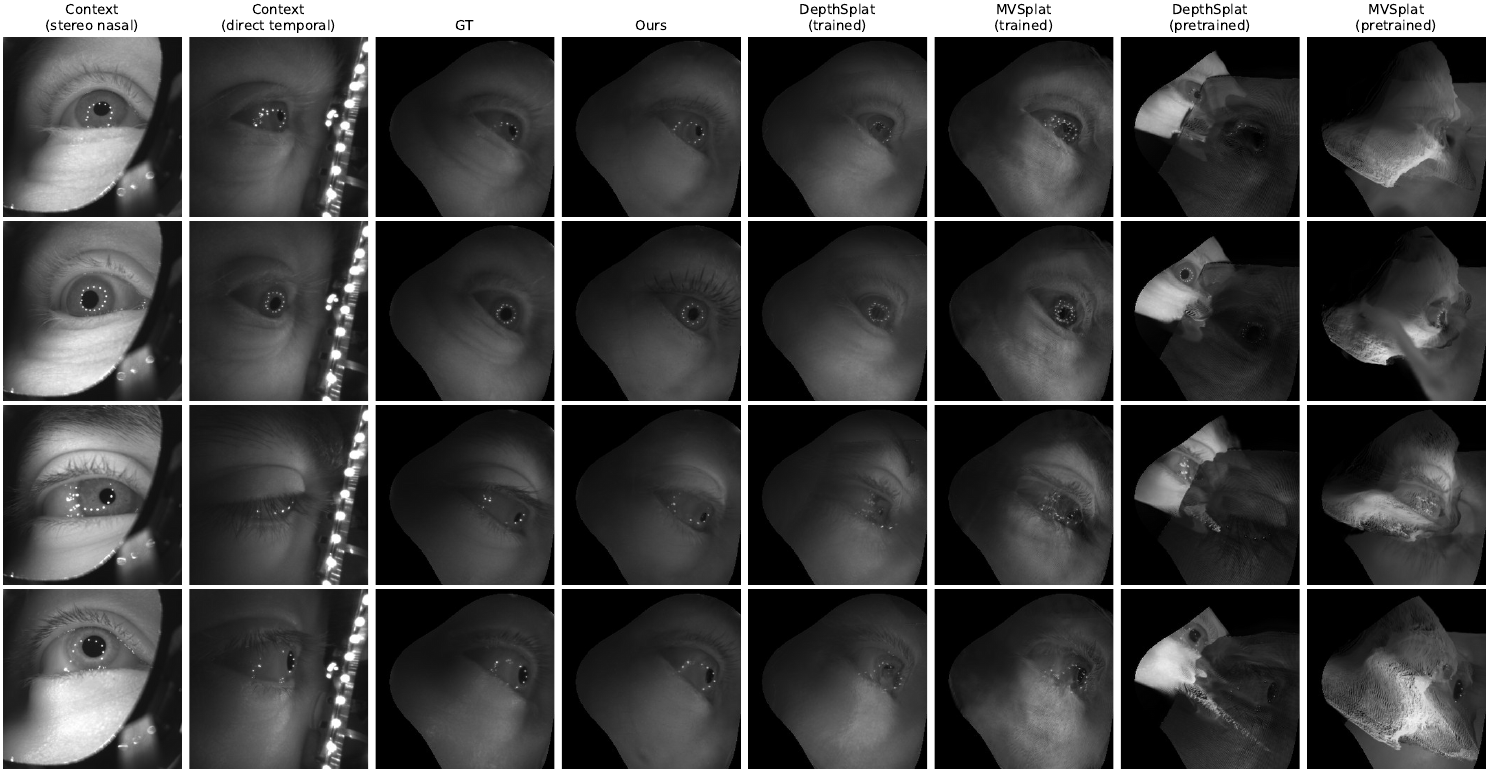}{Novel-view rendering on test-unseen \dsFourV{} subjects from two context views, with MVSplat and DepthSplat trained on \dsFourV{} and with their released checkpoints. Regions hidden by the device housing are masked out.}

We also train the other general methods of Tab.~\ref{tab:nv_merged} on the training splits, MapAnything, MoGe-2 and DA-V2 end to end with their Gaussian heads, the decoder of TRELLIS and a LoRA of ViewCrafter, and evaluate them with the protocol of Tab.~\ref{tab:nv_merged}. The results are shown in  Tab.~\ref{tab:nv_merged_trained} and Fig.~\ref{fig:novel_view_trained}.

\begin{table*}[!t]
    \centering
    \caption{Novel-view rendering on \dsTwoV{}, \dsFourV{} and \dsMasq{}~\cite{Lin2026EyeTracker}, including the general methods trained on our data. The top three methods are baselines for eye modeling, the middle block lists general 3D reconstruction and generation methods with their released checkpoints, and the next block lists the same methods trained on the training splits. $\times$: the method needs at least two context views. --: ControllableGaze is not available on \dsMasq{}, which has no mesh or OCT data.}
    \label{tab:nv_merged_trained}
    \fitbox{\textwidth}{%
    \begin{tabular}{llccccccccccccccc}
    \toprule
     & & \multicolumn{3}{c}{\dsTwoV{}} & \multicolumn{6}{c}{\dsFourV{}} & \multicolumn{6}{c}{\dsMasq{}} \\
    \cmidrule(lr){3-5}
    \cmidrule(lr){6-11}
    \cmidrule(lr){12-17}
     & & \multicolumn{3}{c}{1 view} & \multicolumn{3}{c}{1 view} & \multicolumn{3}{c}{3 views} & \multicolumn{3}{c}{1 view} & \multicolumn{3}{c}{3 views} \\
    \cmidrule(lr){3-5}
    \cmidrule(lr){6-8}
    \cmidrule(lr){9-11}
    \cmidrule(lr){12-14}
    \cmidrule(lr){15-17}
    Method & Source & PSNR $\uparrow$ & SSIM $\uparrow$ & LPIPS $\downarrow$ & PSNR $\uparrow$ & SSIM $\uparrow$ & LPIPS $\downarrow$ & PSNR $\uparrow$ & SSIM $\uparrow$ & LPIPS $\downarrow$ & PSNR $\uparrow$ & SSIM $\uparrow$ & LPIPS $\downarrow$ & PSNR $\uparrow$ & SSIM $\uparrow$ & LPIPS $\downarrow$ \\
    \midrule
    GazePrior~\cite{Dumery2026GazePrior} & OCT-Cam & 20.23 & 0.694 & 0.468 & \textit{27.13} & \textit{0.784} & 0.502 & 27.54 & 0.788 & 0.499 & \textit{16.08} & \textit{0.485} & 0.576 & 17.27 & 0.493 & 0.554 \\
    Lin et al.~\cite{Lin2026EyeTracker} & Static Fit & 14.15 & 0.267 & 0.595 & 11.77 & 0.375 & 0.665 & 15.85 & 0.480 & 0.582 & 11.73 & 0.404 & 0.723 & 14.22 & 0.405 & 0.588 \\
    ControllableGaze~\cite{Li2025ControllableGaze} & OCT-Cam & \textit{21.81} & \textit{0.702} & 0.397 & 19.24 & 0.726 & 0.463 & 18.93 & 0.725 & 0.466 & -- & -- & -- & -- & -- & -- \\
    \midrule
    MapAnything~\cite{Keetha2026MapAnything} (+GS Head) & Pretrained (+OCT-Cam) & 19.62 & 0.653 & 0.355 & 13.66 & 0.575 & 0.500 & 14.50 & 0.681 & 0.481 & 10.73 & 0.353 & 0.806 & 12.32 & 0.451 & 0.758 \\
    MoGe-2~\cite{Wang2025MoGe2} (+GS Head) & Pretrained (+OCT-Cam) & 19.10 & 0.630 & 0.395 & 15.35 & 0.576 & 0.516 & 17.95 & 0.685 & 0.468 & 14.83 & 0.439 & 0.581 & 10.33 & 0.398 & 0.790 \\
    DA-V2~\cite{Yang2024DepthAnythingV2} (+GS Head) & Pretrained (+OCT-Cam) & 18.82 & 0.634 & 0.439 & 14.08 & 0.350 & 0.677 & 15.10 & 0.488 & 0.665 & 10.23 & 0.267 & 0.795 & 12.38 & 0.413 & 0.682 \\
    MVSplat~\cite{Chen2024MVSplat} & Pretrained & $\times$ & $\times$ & $\times$ & $\times$ & $\times$ & $\times$ & 14.34 & 0.490 & 0.621 & $\times$ & $\times$ & $\times$ & 10.68 & 0.102 & 0.746 \\
    DepthSplat~\cite{Xu2025DepthSplat} & Pretrained & $\times$ & $\times$ & $\times$ & $\times$ & $\times$ & $\times$ & 14.56 & 0.406 & 0.664 & $\times$ & $\times$ & $\times$ & 12.15 & 0.417 & 0.767 \\
    TRELLIS~\cite{Xiang2025TRELLIS} & Pretrained & 15.20 & 0.514 & 0.581 & 13.38 & 0.229 & 0.653 & 15.74 & 0.525 & 0.555 & 9.68 & 0.309 & 0.826 & 15.07 & \textit{0.521} & 0.718 \\
    ViewCrafter~\cite{Yu2025ViewCrafter} & Pretrained & 15.56 & 0.501 & 0.514 & 8.07 & 0.439 & 0.711 & 13.14 & 0.514 & 0.676 & 11.95 & 0.298 & 0.658 & 10.90 & 0.323 & 0.775 \\
    \midrule
    MapAnything~\cite{Keetha2026MapAnything} (+GS Head) & OCT-Cam & 20.56 & 0.680 & \textit{0.312} & 13.65 & 0.657 & 0.430 & 15.99 & 0.675 & 0.497 & 11.37 & 0.345 & 0.758 & 12.62 & 0.451 & 0.759 \\
    MoGe-2~\cite{Wang2025MoGe2} (+GS Head) & OCT-Cam & 20.33 & 0.671 & 0.344 & 16.62 & 0.688 & \textit{0.403} & 21.94 & 0.812 & 0.487 & 15.36 & 0.415 & \textit{0.497} & 12.85 & 0.431 & 0.648 \\
    DA-V2~\cite{Yang2024DepthAnythingV2} (+GS Head) & OCT-Cam & 20.37 & 0.679 & 0.331 & 14.27 & 0.653 & 0.466 & 14.54 & 0.615 & 0.566 & 12.41 & 0.391 & 0.728 & 11.95 & 0.431 & 0.710 \\
    MVSplat~\cite{Chen2024MVSplat} & OCT-Cam & $\times$ & $\times$ & $\times$ & $\times$ & $\times$ & $\times$ & \textbf{32.49} & \textit{0.904} & 0.204 & $\times$ & $\times$ & $\times$ & 13.71 & 0.423 & 0.537 \\
    DepthSplat~\cite{Xu2025DepthSplat} & OCT-Cam & $\times$ & $\times$ & $\times$ & $\times$ & $\times$ & $\times$ & \textit{32.00} & \textbf{0.907} & \textit{0.182} & $\times$ & $\times$ & $\times$ & \textit{18.35} & \textbf{0.538} & \textit{0.401} \\
    TRELLIS~\cite{Xiang2025TRELLIS} & OCT-Cam & 15.70 & 0.561 & 0.615 & 19.00 & 0.689 & 0.501 & 19.95 & 0.708 & 0.455 & 11.96 & 0.338 & 0.675 & 14.77 & 0.456 & 0.629 \\
    ViewCrafter~\cite{Yu2025ViewCrafter} & OCT-Cam & 15.53 & 0.480 & 0.482 & 12.64 & 0.504 & 0.677 & 12.99 & 0.582 & 0.658 & 11.80 & 0.309 & 0.637 & 9.07 & 0.181 & 0.708 \\
    \midrule
    \textbf{Ours} & OCT-Cam & \textbf{24.89} & \textbf{0.786} & \textbf{0.128} & \textbf{29.23} & \textbf{0.891} & \textbf{0.137} & 29.42 & 0.894 & \textbf{0.130} & \textbf{18.08} & \textbf{0.502} & \textbf{0.358} & \textbf{19.30} & 0.520 & \textbf{0.324} \\
    \bottomrule
    \end{tabular}%
    }%
\end{table*}

\suppfig{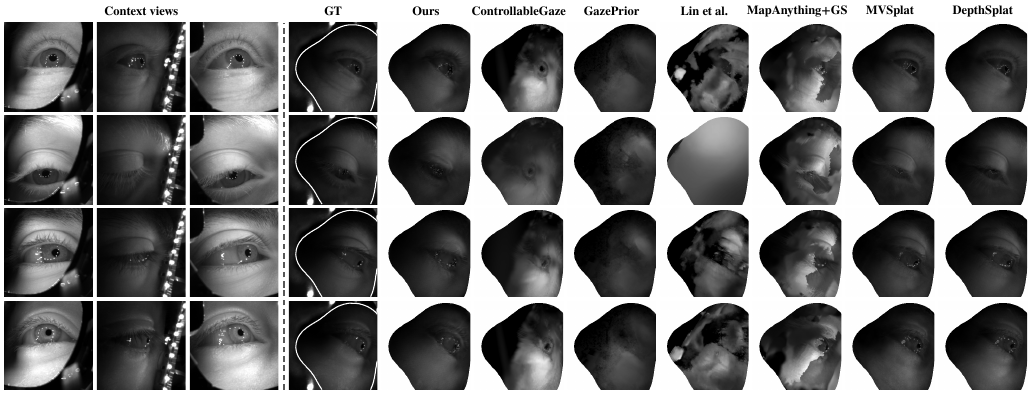}{Novel-view rendering on test-unseen \dsFourV{} subjects from three context views, in the format of Fig.~\ref{fig:qual_nv_main}, with MapAnything+GS, MVSplat and DepthSplat trained on \dsFourV{} (Tab.~\ref{tab:nv_merged_trained}). Three context views (left of the dashed line) are used as inputs to render the target view. Regions hidden by the device housing are masked out.}

\end{document}